\documentclass{ieeeaccess}
\usepackage{cite}
\usepackage{amsmath,amssymb,amsfonts}
\usepackage{algorithmic}
\usepackage{graphicx}
\usepackage{textcomp}
\usepackage{xspace}
\usepackage{comment}

\usepackage[linesnumbered,ruled]{algorithm2e}
\usepackage[font={sf,scriptsize},           
            labelfont={bf,color=accessblue},
            caption=false
            ]                  
           {subfig}

\def\BibTeX{{\rm B\kern-.05em{\sc i\kern-.025em b}\kern-.08em
    T\kern-.1667em\lower.7ex\hbox{E}\kern-.125emX}}
    
\SetArgSty{textup}

\SetCommentSty{mycommfont}

\newcommand{\presets}{\ifmmode \mathcal{P} \else $\mathcal{P}$\xspace \fi}
\newcommand{\preset}[1]{\ifmmode p_{#1} \else $p_{#1}$\xspace \fi}
\newcommand{\config}[1]{\ifmmode c_{#1} \else $c_{#1}$\xspace \fi}
\newcommand{\nal}[1]{\ifmmode \hat{c}_{#1} \else $\hat{c}_{#1}$\xspace \fi}
\newcommand{\losstype}[1]{\ifmmode losstype_{#1} \else $losstype_{#1}$\xspace \fi}
\newcommand{\w}[1]{\ifmmode w_{#1} \else $w_{#1}$\xspace \fi}
\newcommand{\cl}[1]{\ifmmode l_{#1} \else $l_{#1}$\xspace \fi}
\newcommand{\coverage}[1]{\ifmmode \mathbf{C}(#1) \else $\mathbf{C}(#1)$\xspace \fi}
\newcommand{\popcoverage}[1]{\ifmmode \mathbf{PC}({#1}) \else $\mathbf{PC}({#1})$\xspace \fi}

\begin{document}
\history{Date of publication xxxx 00, 0000, date of current version xxxx 00, 0000.}
\doi{10.1109/ACCESS.2017.DOI}

\title{Evaluating and Optimizing Hearing-Aid Self-Fitting Methods using Population Coverage}
\author{
\uppercase{Dhruv Vyas}\authorrefmark{1}, 
\uppercase{Erik Jorgensen\authorrefmark{2}, Yu-Hsiang Wu \authorrefmark{3} and Octav Chipara \authorrefmark{1}}
}
\address[1]{Department of Computer Science, University of Iowa, USA}
\address[2]{Department of Communication Sciences and Disorders, University of Wisconsin-Madison, USA}
\address[3]{Department of Communication Sciences and Disorders, University of Iowa, USA}
\tfootnote{This work is supported by NSF through grants IIS-1838830 and CNS-1750155. 
Additional funding is provided by National Institute on Disability, Independent Living, and Rehabilitation Research (90REGE0013).}

\markboth
{Vyas \headeretal: Evaluating and Optimizing Hearing-Aid Self-Fitting Methods using Population Coverage}
{Vyas \headeretal: Evaluating and Optimizing Hearing-Aid Self-Fitting Methods using Population Coverage}


\begin{abstract}

Adults with mild-to-moderate hearing loss can use over-the-counter hearing aids to treat their hearing loss at a fraction of traditional hearing care costs. These products incorporate self-fitting methods that allow end-users to configure their hearing aids without the help of an audiologist. A self-fitting method helps users configure the gain-frequency responses that control the amplification for each frequency band of the incoming sound. This paper considers how to design effective self-fitting methods and whether we may evaluate certain aspects of their design without resorting to expensive user studies. 
Most existing fitting methods provide various user interfaces to allow users to select a configuration from a predetermined set of presets. We propose a novel metric for evaluating the performance of preset-based approaches by computing their population coverage. The population coverage estimates the fraction of users for which it is possible to find a configuration they prefer. A unique aspect of our approach is a probabilistic model that captures how a user’s unique preferences differ from other users with similar hearing loss. Next, we develop methods for determining presets to maximize population coverage. Exploratory results demonstrate that the proposed algorithms can effectively select a small number of presets that provide higher population coverage than clustering-based approaches. Moreover, we may use our algorithms to configure the number of increments for slider-based methods.

\end{abstract}



\maketitle

\section{Introduction}
\label{sec:introduction}

Hearing loss is an epidemic in the United States; most hearing loss goes untreated. 
The primary treatment for hearing loss is hearing aids (HAs). 
However, of the 48 million Americans with hearing loss, only 14-34\% use HAs \cite{lin_hearing_us_2011, chienPrevalenceHearingAid2012, MarkeTrak10Hearing}. 
A major reason for the low rate of HA adoption is their high cost. 
One study estimated that the average cost of HAs bundled with  several audiologist visits is \$2,500, representing a significant financial expense for 77\% of Americans \cite{jillaHearingAidAffordability2020}. 
Economic barriers to hearing healthcare disproportionately affect minorities; most HA users are affluent, educated, and white \cite{reedTrendsHearingAid2021a, mckeeDeterminantsHearingAid2019, niemanRacialEthnicSocioeconomic2016}. 
Thus, there is a critical need to improve access to hearing care. 
One solution is the advent of over the counter (OTC) HAs. 
The OTC Hearing Aid Act, signed into law in 2017, with subsequent proposed rules delivered in 2021, will enable HAs to be sold over-the-counter without the need for the user to see an audiologist or medical professional. 



The primary function of HAs is to divide the incoming sound into several frequency bands and amplify each band preferentially.
For HAs to address a user's needs it is essential to fit the HA by configuring the gain-frequency response of each band to compensate for the user's hearing loss in that band.
Traditionally, HA fitting is performed by an audiologist, who first measures the user's hearing loss as an \emph{audiogram}. 
Audiologists construct audiograms by presenting pure tone to measure the user's hearing thresholds at frequencies important for speech perception (typically 0.25kHz - 8 kHz). 
A user's hearing loss is characterized by their thresholds across the frequency range relative to the average hearing thresholds of normal-hearing listeners. 
The amount of gain applied in each band is traditionally determined using a prescription formula, commonly NAL-NL2 \cite{Keidser:2011uj}. 

NAL-NL2 is based on theoretical models of speech intelligibility and loudness comfort, as well as empirical data showing differences in gain preferences between different populations (e.g., men vs. women, experienced vs. new HA users). 
Since NAL-NL2 uses theoretical models and population-level statistics, the prescribed NAL-NL2 configuration estimates the average configuration for a sample of users with similar audiograms (and hearing loss).
However, a user's preferred configuration may deviate significantly from their prescribed NAL-NL2 configuration due to individual perceptual, lifestyle, and HA usage factors that differ from person to person \cite{sogaard2019perceptualOticonOn}. 
To customize a user's configuration, a series of visits to the audiologist are generally required to fine-tune the HA's configuration based on their feedback.

OTC HAs reduce cost, time, and other barriers to HA access by shifting the burden of configuring these devices from the audiologist to the end user.
Traditional HA fitting requires sophisticated software, a programming interface, and specialized knowledge typically acquired through years of graduate-level training. 
For the lay user, successful self-fitting must be achieved without any of these helpful resources, which have traditionally been bundled with purchasing a HA. 
To address this limitation, OTC HAs usually provide a user interface that allows users to configure their gain-frequency responses.
Two common strategies are used to design user interfaces -- collection-based and slider-based approaches.
Both approaches operate on a set of predetermined and fixed configurations that we will refer to as \textbf{presets}.
In the case of collection-based methods, the presets are configurations associated with representative types of hearing loss.
In contrast, slider-based methods directly manipulate various aspects of configurations. For example, a loudness (i.e., overall amplitude) slider increases the overall gain. The presets are the union of all the possible configurations that the controllers may reach. It is common for collection-based methods to include a small number of presets, whereas slider-based methods typically include more presets. We provide additional details about these two methods, including concrete examples of their use with existing HAs in Section \ref{sec:related-work}.

A key challenge associated with advancing the OTC fitting methods is that their design requires extensive and costly user studies.
Similar challenges also occur when comparing different approaches.
These questions will become increasingly important as more OTC HAs are commercialized. 
Therefore, we pose the question of whether it is possible to create metrics that we can use to assess OTC HA fitting methods without resorting to extensive user studies.
We propose a new metric -- \textbf{population coverage} -- to evaluate, compare, and optimize fitting methods.
The population coverage estimates how well a set of presets \presets meets the needs of users with mild-to-moderate hearing loss, the population that would benefit from an OTC HA the most.
We will develop statistical models to estimate the population coverage of a general self-fitting method given (1) statistics regarding the typical hearing loss of users in a population of interest and (2) the set of presets used in the fitting methods.
Moreover, we will show that it is possible to optimize various parameters of a fitting method to maximize population coverage.

Our work builds on a line of research in audiology that aims to identify representative audiograms.
Ciletti and Flamme \cite{ciletti2008prevalence} used audiograms provided from two publicly available datasets and applied cluster analysis to identify audiograms representative of the US population.
The audiograms within a cluster were more similar than those in different clusters.
The clustering analysis results may be used to generate a set of presets by converting the centroid audiogram of each cluster to a REAR configuration using NAL-NL2. 
A similar approach to generating presets was proposed in \cite{jensenCommonConfigurationsRealEar2020}.

We extend these approaches in two significant directions.
First, the above approaches assume that NAL-NL2 accurately predicts a user's most preferred configuration.
Empirical evidence shows that this assumption is incorrect and yields suboptimal configurations to be selected as presets.
We develop a statistical model that characterizes how a user's preference may deviate from their NAL-NL2 prescription (see Section \ref{sec:coverage}).
Second, our techniques are more general in that we can use them for both collection- and slider-based approach. 
We present algorithms to selected presets for both collection- and slider-based approaches to maximize the population coverage.
Finally, our results show our approach's superiority over standard cluster-based techniques in constructing presets.

The remainder of the paper is organized as follows.
Section \ref{sec:related-work} summarizes prior work on self-fitting methods.
A formal definition of population coverage and methods for computing it for preset-based self-fitting methods is described in Section \ref{sec:coverage}.
Algorithms to optimize the presets for collection- and slider-based approaches are included in Section \ref{sec:preset-optimization}.
Results comparing different methods for generating presets are provided in Section \ref{sec:experiments}, and their importance is discussed in Section \ref{sec:discussion}.
Section \ref{sec:conclusions} provides conclusions and discusses future work.

\section{Related Work}
\label{sec:related-work}


\noindent
\textbf{Slider-based approaches:}
A common approach to allow users to personalize OTC HAs is to use sliders (or wheels) that manipulate the gain-frequency responses HAs used to amplify each channel.
For example, the Ear Machine approach presents users with two wheels, one which sliders loudness and one which controls fine tuning \cite{nelsonSelfAdjustedAmplificationParameters2018}. 
The loudness slider enables users to vary the overall gain and compression parameters of the HA and the fine-tuning wheel enables users to vary the tilt of the gain-frequency response around a 2 kHz fulcrum. 
In another approach, called Goldilocks, the user is presented with three parameters they are able to adjust using up and down arrows: fullness (low frequency cut), crispness (high frequency boost) and loudness (overall amplification) \cite{boothroydGoldilocksApproachHearingAid2017, mackersieGoldilocksApproachHearing2019}. 
The user first adjusts loudness, then crispness, then loudness again, then fullness, then any parameter until they find a gain-frequency response that is ``just right.'' 

\noindent
\textbf{Collection-based approaches:}
These approaches can lead to successful self-fitting \cite{boothroydAmplificationSelfAdjustmentControls2022, brodyComparisonPersonalSound2018}. 
However, because these methods require the user to manipulate various parameters on continuous scales, the self-fitting process can be cumbersome and cognitively challenging. 
For example, six of the 26 users in the evaluation of the Goldilocks required help with the interface and there was a wide range in time to completion among users, indicating that some users likely found the procedure more difficult than others \cite{boothroydGoldilocksApproachHearingAid2017}. 
Recent MarkeTrak survey data found up to half of potential OTC users would not be comfortable tuning their own HAs \cite{edwardsEmergingTechnologiesMarket2020a}. 
One way to simplify the self-fitting process is to use a unordered collection of presets \cite{urbanskiNewEvidenceBasedFitting2021, sabinValidationSelfFittingMethod2020}. 
In a collection approach, gain-frequency responses that can accommodate most users are predetermined. 
The self-fitting process then involves arriving at the user's most  preferred preset. 
Presumably, the preset approach can both shorten the self-fitting process and make the procedure easier, reducing failures during the self-fitting process. 
Although some studies have investigated the use of presets in OTC HAs, it remains an open problem how to determine the right set of presets and what is the best strategy of identifying user's preferred preset in an easy and efficient manner been established. 
In this paper, we focus the former problem while not directly addressing the latter.  

In \cite{urbanskiNewEvidenceBasedFitting2021}, presets were derived using audiograms of adults with mild-to-moderate hearing loss, taken from a large national database. 
First, they found all possible gain-frequency responses based on the audiogram set. 
Then, they found the four presets that fit the largest number of NAL-NL2 targets within $\pm$ 5 dB from 0.25 - 4 kHz. 
The four presets could match NAL-NL2 targets within 10 dB for 70\% of the audiograms in the database. 
The presets were then empirically tested by first assigning different presets to participants using various fitting methods and having participants complete speech testing using the presets selected in each method. 
Presets were determined either based on the participant's audiologist-administered audiogram, a self-test audiogram, listening to the different presets in quiet and in noise, a questionnaire, or random assignment. 
The presets determined by each selection method were generally different from one another.
However, none of the fitting methods besides the questionnaire and random assignment methods resulted in speech perception scores that differed significantly from those obtained when participants were individually fit to NAL-NL2 targets. 
Taken together, the findings from that study suggest that the process by which presets are determined for individual listeners impacts what preset is chosen--but several different presets can yield comparable efficacy for speech perception for any given listener. 
\section{Coverage of Self-fitting Methods}
\label{sec:coverage}

In this section, we consider the problem of estimating the fraction of users with mild-to-moderate hearing loss whose hearing needs can be met by a set of $N$ presets \presets (i.e., $|\presets| = N$).
To do so, we need to answer three questions:
\begin{enumerate}
    \item How do we define the subset of users with mild-to-moderate hearing loss?
    \item When are the hearing needs of a user met by a preset?
    \item How do we estimate the fraction of users of interest that are covered by a set of presets? 
\end{enumerate}
\noindent We will answer each of these questions in the following subsections.

\subsection{Target Population}
\label{subsec:audiogram-selection}

The population of users that would benefit the most from OTC HAs include those who have perceived mild-to-moderate hearing loss.
The National Health and Nutrition Examination Surveys (NHANES) \cite{nhanes_dataset} from 1996-2016 includes audiograms characterizing common hearing loss configurations.
For each user $u$, NHANES includes a \textbf{population weight} \w{u} representing the prevalence of that type of hearing loss in the population.
We focus on a subset of older individuals with mild to moderate sensorineural hearing loss from  NHANES who meet the following criteria:
1) they are between 55 and 85 years old (inclusive); 
2) their audiometric thresholds from .25 kHz to 6 kHz less than or equal to 75 dB HL, with a threshold at 8 kHz that did not exceed 120 dB HL; 
3) a four-frequency puretone average (.5, 1, 2, and 4 kHz) greater than 20 dB HL but less than 50 dB HL (World Health Organization of mild-to-moderate hearing loss, \cite{olusanya2019hearingWHO}); 
4) normal middle ear function defined as peak tympanometric peak pressure greater than or equal to -50 daPa and less than or equal to 50 daPa, ear canal volume equal to or greater than 0.5 ml but less than or equal to 2 ml, and compliance equal to or greater than 0.3 ml and less than or equal to 1.5 ml. If only one ear qualified, only that ear was included in the dataset. 
Finally, all audiometric and tympanometric data had to be complete with no missing values. 
The final selected subset included 1979 audiograms (mean age=67.77 years, SD=8.02 years; 1018 female, 961 male; 813 bilateral, 1166 unilateral).

For each user $u$, we compute the prescribed \textbf{NAL-NL2 configuration} \nal{u}.
The NAL-NL2 configuration is computed using the NAL-NL2 software \cite{nalnl2_software}. 
whose parameter settings were configured similar to \cite{urbanskiNewEvidenceBasedFitting2021} and \cite{sabinValidationSelfFittingMethod2020}: 
1) thresholds entered as air conduction; 
2) broadband input signal level of 65 dB SPL; 
3) 18-channels with multichannel compression limiting; 
4) default directionality (0 degrees azimuth); 
5) default microphone (head surface); 
6) experienced HA users;
7) compression speed to dual;
8) non-tonal language. 
The user's gender and age are set according to the NHANES data. 
NHANES dataset includes audiograms for unilateral hearing loss (hearing loss in one ear) or bilateral hearing loss (hearing loss in both ears).
Bilateral hearing loss could be treated by fitting one or both of the ears.
For bilateral audiograms, we compute four sets of NAL-NL2 configurations: unilateral left, unilateral right, bilateral left, and bilateral right. 
Thus, we compute a total of 4418 NAL-NL2 configurations.

\subsection{Population Coverage}

A configuration $c$ is a six-dimensional vector ($\config{i} \in \mathbb{R}^{6}$) representing the Real-Ear-Insertion-Gains (REIGs) at frequencies 500, 1000, 2000, 3000, 4000, and 6000 Hz.
The notation $c[f]$ ($f \in \{500, 1000, 2000, 3000, 4000, 6000\}$) refers to the gain associated with frequency $f$ in the vector $c$.
We will select a subset of $N$ of configurations as presets.
We define the \textbf{preset coverage} of a preset \preset{i} to include all the configurations $c$ located within the ball centered at \preset{i} and with a radius $R$.
The distance between a configuration $c$ and \preset{i} is the maximum absolute difference computed across the six frequencies:

\begin{equation}
    B_R(\preset{i}) = \{c \in \mathbb{R}^{6}~|~\max |c - \preset{i}| \le R\}
\end{equation}
\noindent We assume that all the configurations within the ball are covered by preset \preset{i}. 
Audiologists are typically satisfied to find a configuration within $\pm~5~\textrm{dB}$, so we set $R = 5~\textrm{dB}$.

The coverage of all presets in \presets is $C(\presets)$:
\begin{equation}
\coverage{\presets} = \bigcup_{\preset{i} \in \presets} B_R(\preset{i})
\end{equation}

\noindent Ideally, the presets included in \presets maximize the fraction of users whose configurations are included in $C(\presets)$.

Next, we consider determining whether a user’s configuration is covered. 
A naive approach would be to assume that a user's $u$ preferred configuration coincides with that prescribed by NAL-NL2 (as done in prior work \cite{urbanskiNewEvidenceBasedFitting2021, jensenCommonConfigurationsRealEar2020}).
Consistent with this assumption, a user would be covered if its prescribed NAL-NL2 configuration \nal{u} would be included in $C(\presets)$.
Then, the fraction of the population covered would be the sum of the population weights of covered users.

Emerging empirical evidence shows, however, that a user's preferred configuration often deviates from their NAL-NL2 prescription.
In the following, we will propose a statistical model that characterizes how the users' preferences may deviate from the NAL-NL2 model. 
We start by observing that the average of the NAL-NL2 prescriptions of a group of users with similar audiograms (and hearing losses). 
However, there is a significant inter-subject variability due to individualized preferences.
Moreover, empirical evidence shows that a user's preferred configuration often differs from the NAL-NL2 configuration either its ``overall gain'' and/or ''slope'' \cite{}.
Therefore, the goal is to model how a user's preferred configuration may deviate from the NAL-NL2 prescription in overall gain and slope.

\begin{figure*} 
\centering
\subfloat[Transfer functions]{\includegraphics[width=0.32\linewidth]{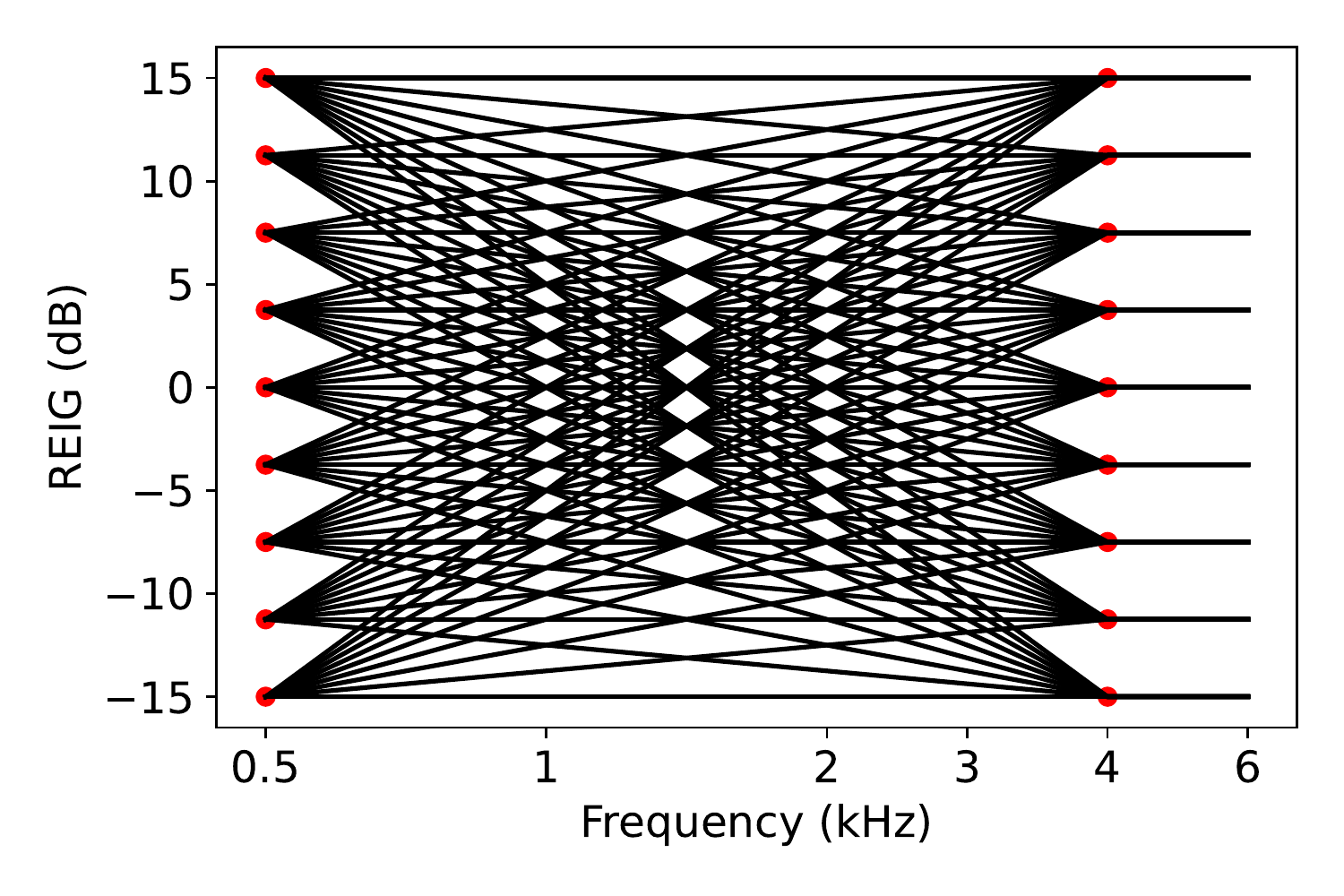} \label{fig:transfer-functions}}
\hfil
\subfloat[Example of an REIG from NHANES]{\includegraphics[width=0.32\linewidth]{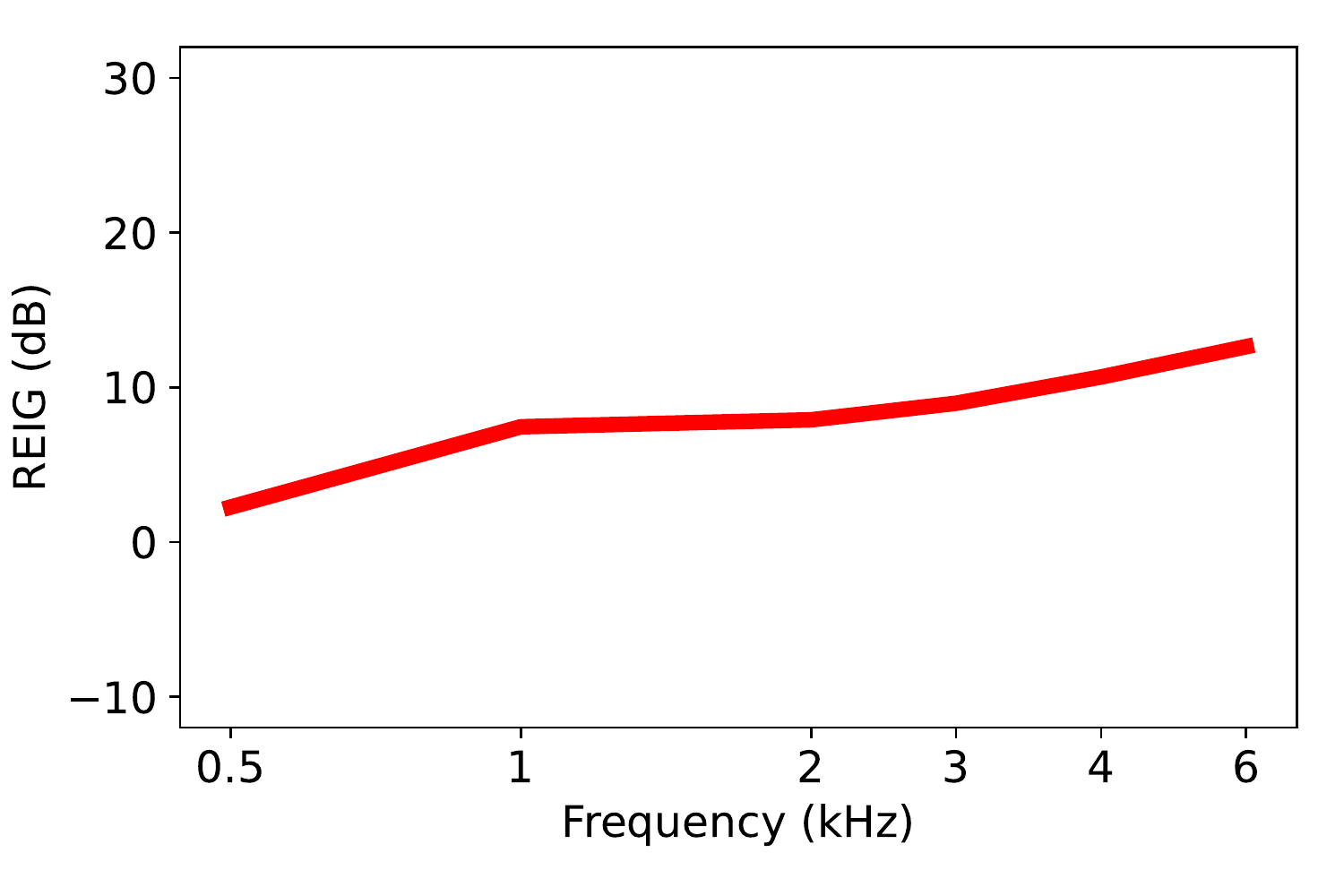} \label{fig:nhanes-reig-example}}
\hfil
\subfloat[Transfer functions superimposed on the REIG to create REIG variations]{\includegraphics[width=0.32\linewidth]{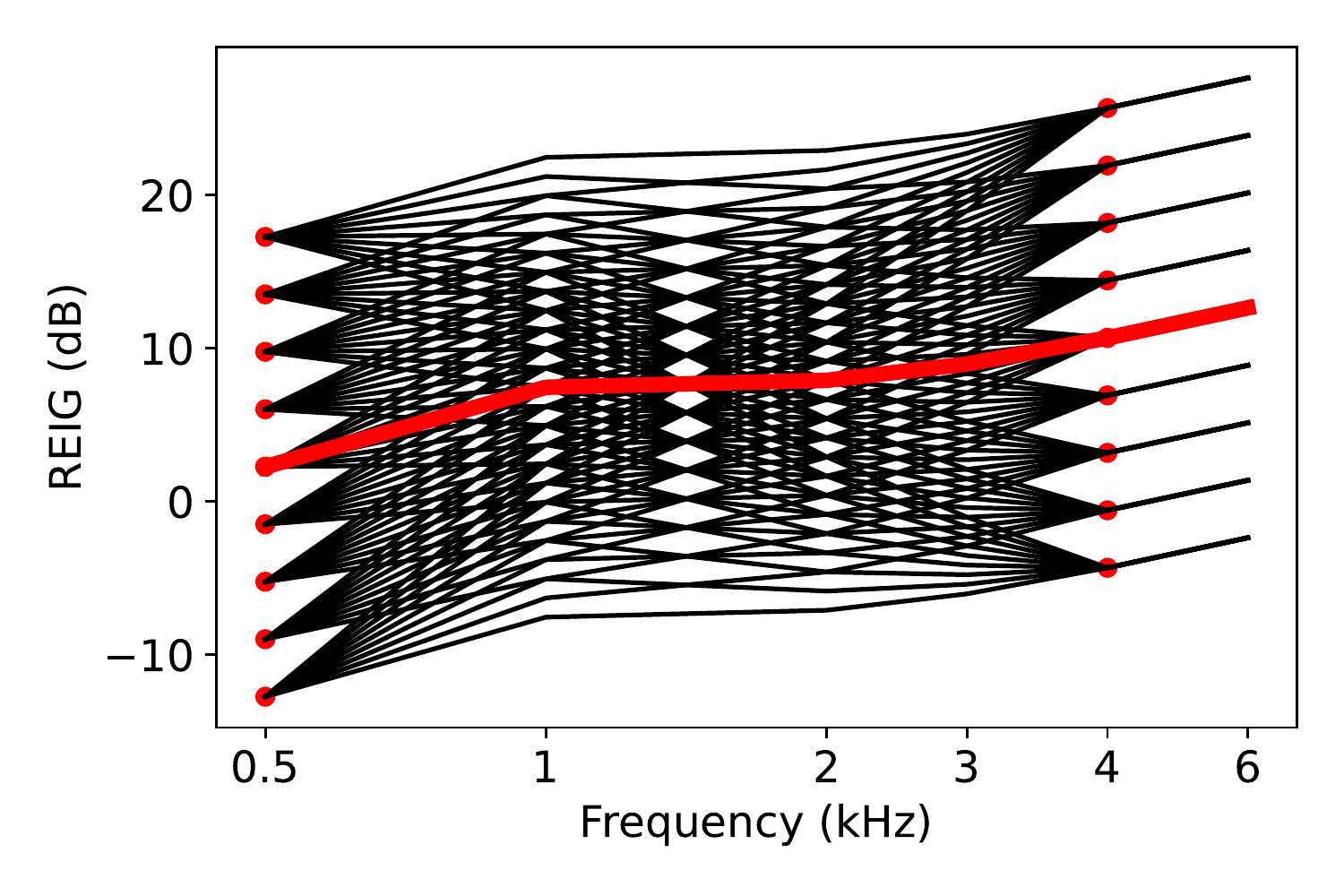}\label{fig:variations}}
\caption{Example of an REIG from NHANES and its derived variations from transfer functions}
\label{fig:nhanes-reig-variations-combined}
\end{figure*}

To this end,  we introduce \textbf{transfer functions} (see Figure \ref{fig:transfer-functions}).
A transfer function is a log-linear function ($T_j : \mathbb{R} \rightarrow \mathbb{R}$) describing how a user's $u$ preferred configuration \config{u} may deviate from their NAL-NL2 configuration \nal{u} at $f$ Hz.

\begin{equation}
    \config{u,j}[f] = \nal{u}[f] + T_j(f)
\end{equation}

\noindent We capture a user's $u$ potential \textbf{preferred configurations} \config{u,j} using several transfer functions $T_j$, each yielding changes in overall gain and slope.

We construct the transfer functions as follows.
Empirical evidence suggests that the magnitude of the deviations at 500 and 4000 Hz are within $\pm$ 15 dB~\cite{mackersieHearingAidSelfAdjustment2020,boothroydAmplificationSelfAdjustmentControls2022}.
We divide the range from -15 dB to 15 dB in increments of 3.75 dB and create an anchor point at each increment at frequencies of 500 and 4000 Hz  (see Figure \ref{fig:transfer-functions}).
We selected increments of 3.75 dB as individuals cannot perceive the difference between configurations with smaller REIG differences~\cite{caswellmidwinterDiscriminationGainIncrements2019}.
Then, each transfer function is constructed by fitting a line for each pair of anchor points. 
By connecting each pair of anchor points from 500 to 4000 Hz, we create 81 possible transfer functions. 
Accordingly, for a user $u$, there are 81 possible preferred configurations, 80 of which differ from the user's NAL-NL2 configuration.
For example, Figure \ref{fig:nhanes-reig-example} plots the NAL-NL2 prescription for one of the users in NHANES.
Figure \ref{fig:variations} plots the user’s preferred configurations computed by adding each transfer function shown (shown in Figure \ref{fig:transfer-functions}) to the user's NAL-NL2 configurations (shown in Figure \ref{fig:nhanes-reig-example}).

The transfer functions are designed to cover the range of possible deviations from NAL-NL2.
However, not all potential preferred configurations are equally likely --- configurations that are closer to NAL-NL2 tend to be more likely.
With each of the 81 possible configurations, we associate a likelihood that is determined by a 2D Gaussian parameterized by average deviation in the low (500 and 1000 Hz) and high frequencies (2000, 3000, and 4000 Hz).
Studies \cite{mackersieHearingAidSelfAdjustment2020} and \cite{boothroydAmplificationSelfAdjustmentControls2022} collected statistics about the deviations between the configuration prescribed by NAL-NL2 and those preferred by users. 
Using this data, we empirically determine the mean and standard deviation of the 2D Gaussian (see Figure \ref{fig:gauss}). 
The weight \cl{u,j} of each potential preferred configuration is \config{u,j} determined according to the fitted 2D Gaussian.

It is important to note that the NHANES dataset comprises users with two types of hearing loss -- unilateral loss (hearing loss in one ear) and bilateral loss (hearing loss in both ears).
Users with bilateral hearing loss can be fit in 3 different ways.
1) Unilateral fitting for left ear.
2) Unilateral fitting for right ear.
3) Bilateral fitting for both ears.
We therefore have four different NAL-NL2 configurations (\nal{u(uni;left)}, \nal{u(uni;right)}, \nal{u(bi;left)} and \nal{u(bi;right)}) for bilateral hearing loss users.

We define a user with unilateral hearing loss $u$ to be covered if the sum of weights of the covered potential preferred configurations exceeds a threshold ($\gamma$).
A preferred configuration \config{u,j} is covered if it is included in $C(\presets)$.
We define a user with bilateral hearing loss $u$ to be covered if the sum of weights of the covered potential preferred configurations exceeds a threshold ($\gamma$) for all four configurations.
Preferred configurations \config{u(uni;left), j}, \config{u(uni;right), j}, \config{u(bi;left), j} and \config{u(bi;right), j} are covered if they are included in $C(\presets)$).
The \textbf{population coverage} is then the sum of the population weights of the covered users.
The details of the coverage computation are included in Algorithm \ref{alg:coverage-computation}.

\begin{algorithm}[t]
\small
\DontPrintSemicolon

    \SetKwInOut{Input}{Input}
    \SetKwInOut{Output}{Output}
    \SetKwRepeat{Do}{do}{while}%
    \Input{\presets - presets\\
          U - set of users \\
          \losstype{u} - (0 = unilateral loss, 1 = bilateral loss) \\
          \w{u} - population weights \\
          \nal{u} - NAL-NL2 prescription for user $u$ \\
          $T_j$ - the j-th transfer function \\
          $\cl{u,j}$ - likelihood of potential config. \config{u,j} \\
          $\gamma$ - user coverage threshold ($\gamma = 0.8$)
        }
         
    \Output{\popcoverage{\presets} -- population coverage}
    \BlankLine
    \popcoverage{\presets} = 0 \;
    \For{$u \in U$} {
        \eIf{\losstype{u} == 0} {
            $\textrm{user\_coverage} = 0$ \;
            \For{$j=\overline{1,81}$} {
                \tcc{Compute the user's potential config.}
                $\config{u,j} = \nal{u} + T_j$ \;
                \If{$\config{u, j} \in \coverage{\presets}$} {
                    user\_coverage = user\_coverage + \cl{u,j}
                }
            }
            \BlankLine
            \If{$\textrm{user\_coverage} \ge \gamma$} {
                \tcc{User is covered}
                $\popcoverage{\presets} = \popcoverage{\presets} + \w{u}$ \;
            }
        }
        {
            \ForEach{$fit\_type \in \{(uni;left), (uni;right), (bi;left), (bi;right)\}$} 
            {
                $\textrm{user\_coverage}_{fit\_type} = 0$ \;
                \For{$j=\overline{1,81}$} {
                    \tcc{Compute the user's potential config.}
                    $\config{u(fit\_type),j} = \nal{u(fit\_type)} + T_j$ \;
                    \If{$\config{u(fit\_type), j} \in \coverage{\presets}$} 
                    {
                       $\textrm{user\_coverage}_{fit\_type} = \textrm{user\_coverage}_{fit\_type} + \cl{u,j}$ \;
                    }
                }
            }
            \BlankLine
            \If{$\textrm{user\_coverage}_{uni;left} \ge \gamma$ \& $\textrm{user\_coverage}_{uni;right} \ge \gamma$ \& $\textrm{user\_coverage}_{bi;left} \ge \gamma$
            \& $\textrm{user\_coverage}_{bi;right} \ge \gamma$} {
                \tcc{User is covered}
                $\popcoverage{\presets} = \popcoverage{\presets} + \w{u}$ \;
            }
        }
    }
    \caption{Compute population coverage of presets \presets}
        \label{alg:coverage-computation}

\end{algorithm}

\Figure[t!](topskip=0pt, botskip=0pt, midskip=0pt)[scale=0.5]{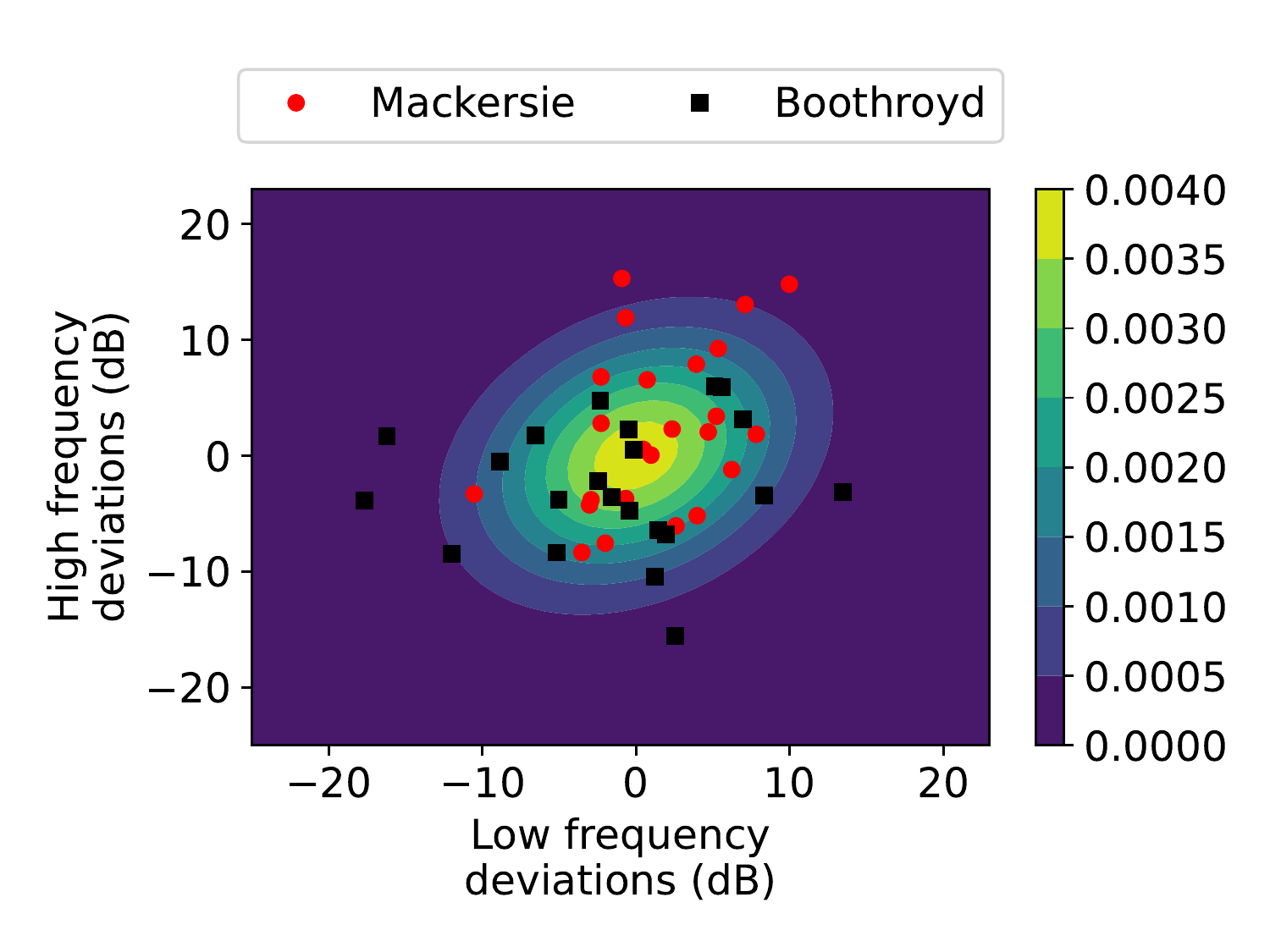}
{2D Gaussian plot of transfer function weights with data points from literature \label{fig:gauss}}

\subsection{Example}
\label{subsec:coverage-calculation}



\Figure[t!](topskip=0pt, botskip=0pt, midskip=0pt)[scale=0.5]{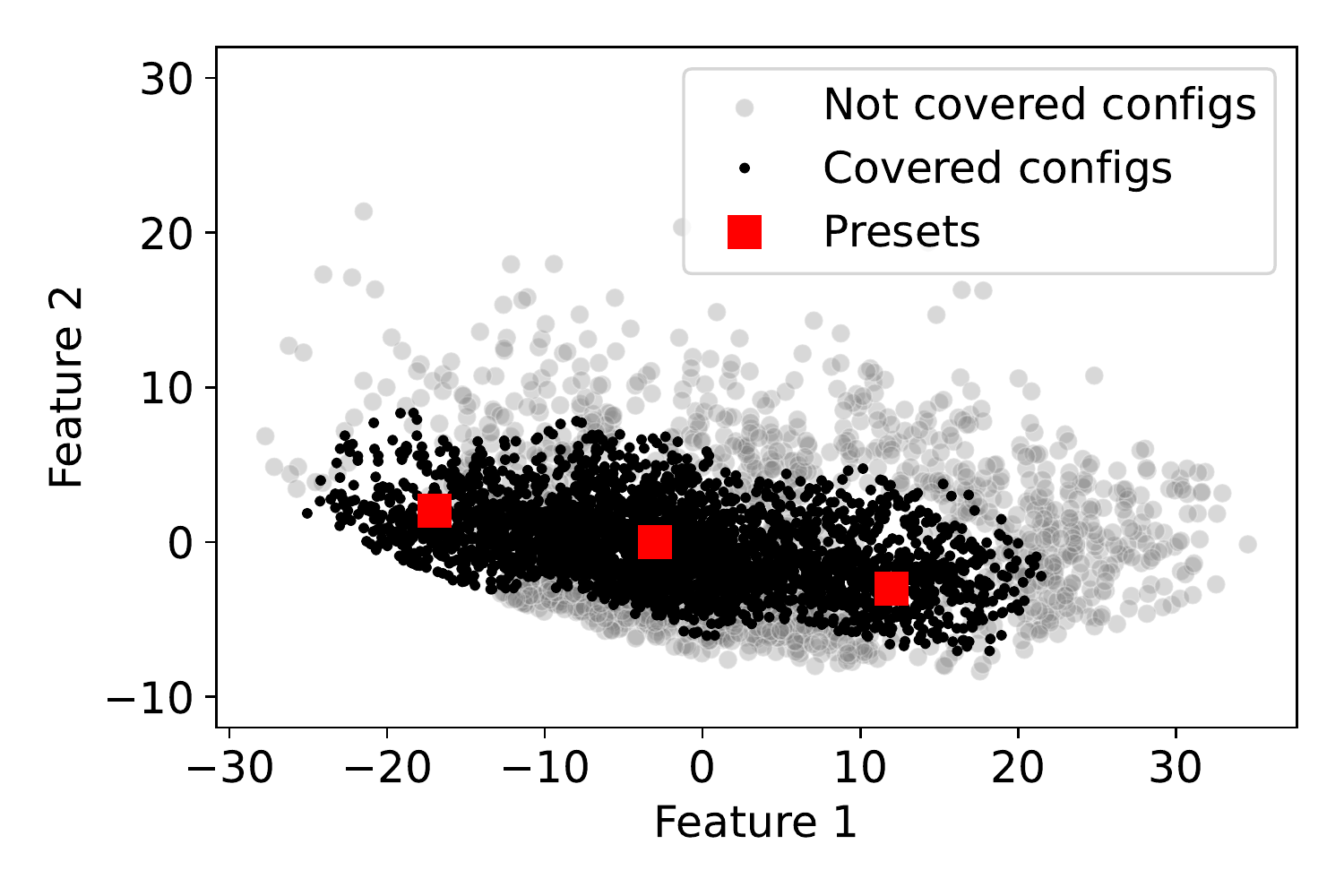}
{Coverage example. Red squares represent example presets in 2D PCA space. Black dots represent the NHANES configurations covered by these distinct presets. Grey dots represent NHANES configurations that are not covered by any of the presets. \label{fig:coverage_example_pca}}

To illustrate how population coverage works, we represent example presets and NHANES user configurations covered by them in PCA space (Shown in Figure \ref{fig:coverage_example_pca}).
PCA space is obtained by conducting Principal Component Analysis on configurations obtained from NHANES data.
There are three example presets represented with bright red squares.
Black dots are the NHANES configurations covered by these distinct presets.
Grey dots represent the NHANES configurations not covered by any of these example presets.
Some of the covered configurations can be covered by multiple presets.

\section{Preset Optimization}
\label{sec:preset-optimization}

In this section, we focus on the problem of fitting OTC HAs by either using preset collection- or slider-based approaches.
In both cases, the problem is to balance the population coverage of the \presets and the number of presets included.
The number of presets included is a proxy for the complexity of the user interface provided to the end user and the increased time to identify the user's preferred configuration.
Generally, user interfaces that expose fewer presets are likely easier to use but provide reduced population coverage.
Conversely, increasing the number of presets increases population coverage at the cost of a more complex user interface.
The algorithms discussed in this section can evaluate the trade-off between population coverage and the number of presets.
Researchers may use this information to evaluate the effectiveness of existing self-fitting methods or to identify promising new approaches.


\subsection{Collection-based Self-fitting Methods}


\Figure[t!](topskip=0pt, botskip=0pt, midskip=0pt)[scale=0.5]{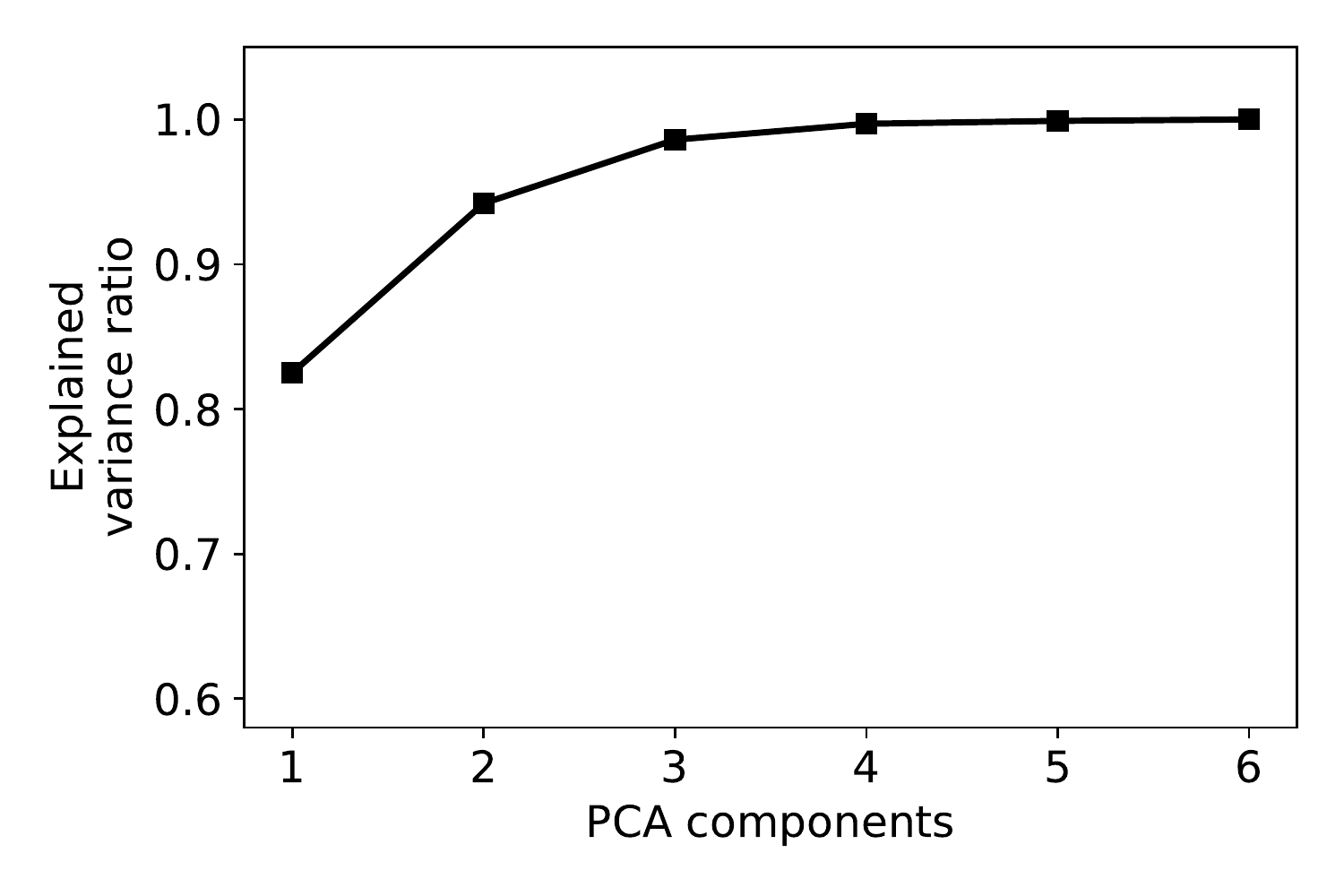}
{PCA explained variance ratio for different number of components. The graph indicats that a large fraction of the variance may be explained using 2 -- 3 dimensions.  \label{fig:pca_explained_variance}}


\Figure[t!](topskip=0pt, botskip=0pt, midskip=0pt)[scale=0.5]{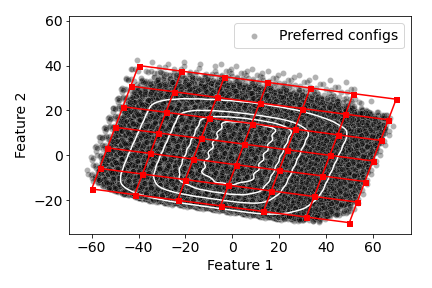}
{Overlaying a grid in the 2 dimensional PCA space \label{fig:pca_with_grid}}

Collection-based self-fitting methods maximize the population coverage of their presets.
The input to the optimization problem is the sample of users with mild-to-moderate hearing loss, each with an associated population weight $\w{u}$.
A feasible solution \presets must include $N$ configurations (i.e., $|\presets| = N$).

A key challenge to determining \presets is the high dimensionality of the configuration space.
To simplify the problem, we use PCA to reduce the dimensionality of the configuration space to two dimensions.
Figure \ref{fig:pca_explained_variance} plots the fraction of the explained variance as the number of dimensions is increased from 1 to 6.
The PCA transformation is applied for all the NAL-NL2 prescriptions of the selected users.
The figure indicates that using two dimensions accounts for 95\% of the total variance.
Therefore, it is possible to map the high-dimensional configuration space into two dimensions with little accuracy loss.
This result can be explained by a significant correlation between the frequency-gain responses across frequencies.

We can use the reduced two-dimensional space to simplify the original optimization problem.
We observe that it is possible to construct a two-dimensional grid with $G$ vertices that spans the space of possible configurations, as shown in Figure \ref{fig:pca_with_grid}.
Then, a solution to the optimization problem involves selecting $N$ out of the $G$ configurations in the grid.
Note that the granularity of the grid influences the population coverage of the selected presets (\presets).
For course-grained grids (i.e., when $G$ is small), we can identify the best solution by evaluating the $C_{N}^{G}$ possible combinations and selecting the one which provides the most population coverage.
In contrast, it is computationally expensive to brute-force all solutions for finer-grained grids (i.e., when $G$ is large), and more efficient algorithms are necessary.

We propose two approaches to solve the problem in the case of fine-grained grids.
A simple strategy is to use a greedy algorithm that iteratively adds to \presets the preset that improves the population coverage the most.
The algorithm maintains a set $X$ of candidate configurations on the 2D grid that can be selected as presets.
Additionally, the set $\presets'$ contains the configurations already selected to be presets.
Initially, $\presets' = \emptyset$ and $X$ includes all the points in the grid.
For each configuration $\config{}$ ($\config{} \in X$), the algorithm computes population coverage of $\presets' \cup \{\config{} \}$.
Then, the configuration that produces the largest increase in population coverage is added to $\presets'$ and removed from $X$.
While the greedy algorithm is computationally efficient, the greedy choice does not always identify the best set of presets.

A better alternative is to use a genetic algorithm (GA) to find the set of \presets.
We encode an individual's chromosome as a bit vector whose size equals the number of points on the two-dimensional grid.
If a grid point is selected to be a preset, then its associated bit is set to one; otherwise, the bit is set to zero.
The GA starts with an initial population containing individuals generated by setting N bits to equal 1 at random locations in the chromosome.
The fitness of each individual is determined by its population coverage.

The next generation of individuals is constructed as follows. First, the individual with the highest fitness is added to the next generation. Then, half of the next generation is generated using cross-over. Cross-over involves two individuals with fitness in the top 50\% of the population as parents. Then, a random location, $L$, is selected. A new individual is created by copying the first L bits from one parent and the remaining $N-L$ bits from the other parent. The remaining 50\% of the population is obtained by mutating the chromosomes of individuals. Specifically, we select a random individual and mutate its chromosome: we flip two bits, one whose value was originally one and one whose value was originally zero. This operation enforces that the number of ones in the chromosome still sum up to $N$. The algorithm terminates after a predetermined number of iterations (set to 500 iteration in our experiments).

\subsection{Slider-Based Self-fitting Methods}

\begin{figure}[h!]
\centering
\vfill
\subfloat[Slider user interface]{\includegraphics[width=0.5\linewidth]{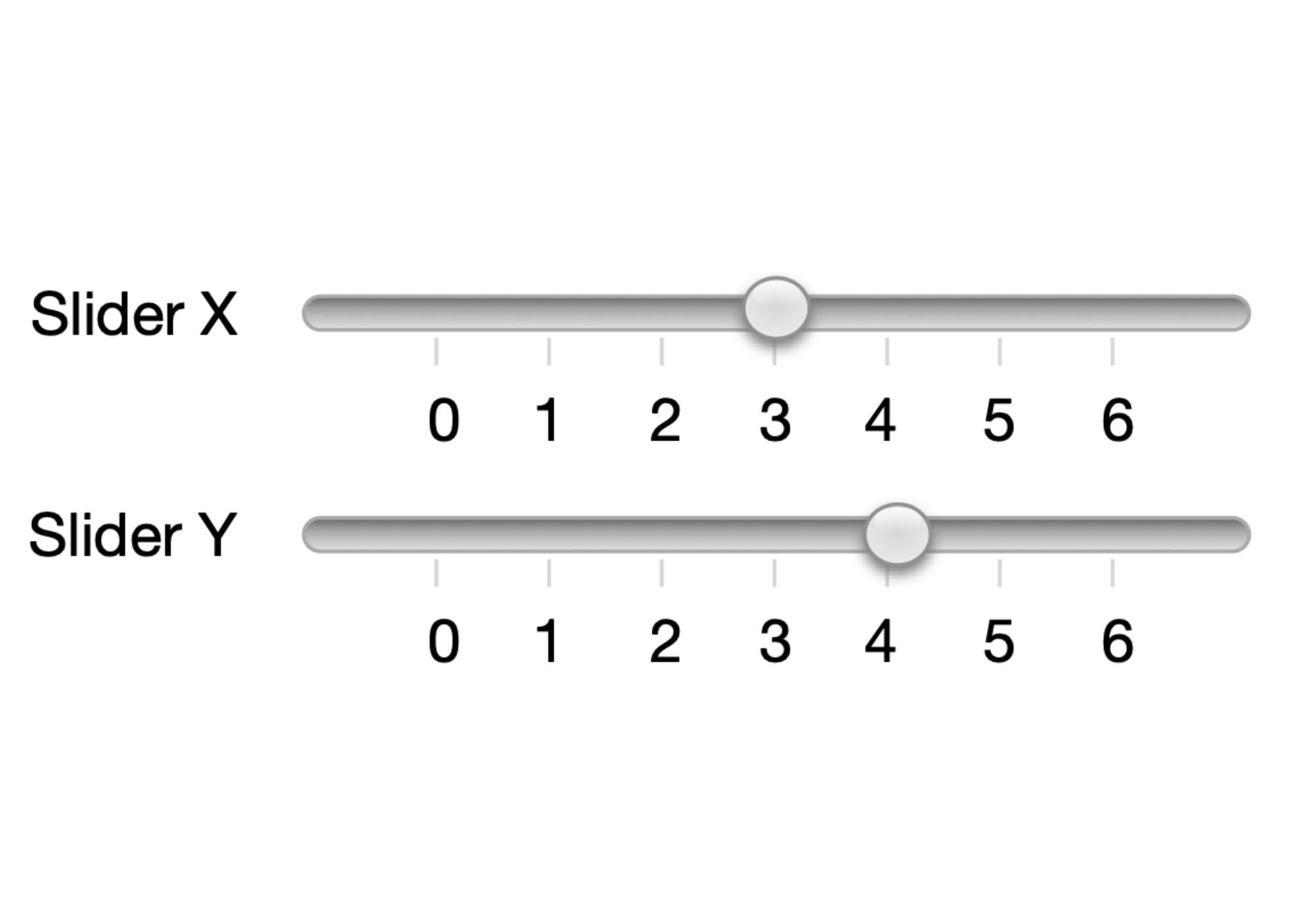}\label{fig:sliders:ui}}
\hfill
\subfloat[Controller grid]{\includegraphics[width=0.45\linewidth]{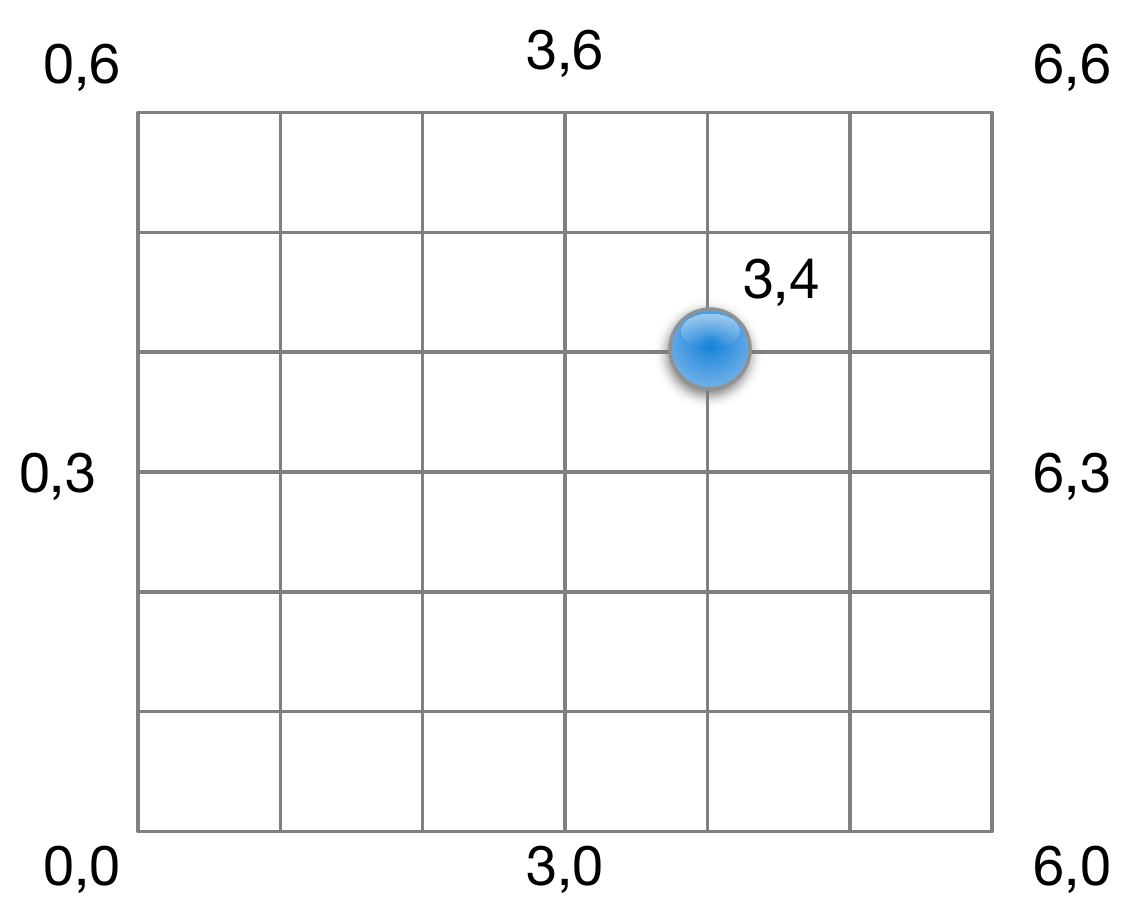} \label{fig:sliders:grid}}
\caption{A slider interface may be used to selected configuration. The controller grid is shown in Figure \ref{fig:sliders:grid} and the current configuration is indicated by the blue circle.}
\label{fig:grid}
\end{figure}

Another common approach to configuring OTC HAs is providing a user interface with one to three sliders.
An example of an interface with two sliders is shown in Figure \ref{fig:sliders:ui}.
An intuitive way of thinking about how the user selects the configuration is to image a ball representing the current configuration selection. The ball moves when the sliders are manipulated. The ball's movement is constrained to occur on a two-dimensional grid which we refer to as the controller grid. As expected, the slider ``x'' controls the ball's horizontal position, whereas the ``y'' controller controls its vertical position. The open question is how to map the ball's position in the controller space to a configuration in the space of possible configuration.

We will create a simple isomorphic mapping between the controller and configuration spaces. We start by identifying a bounding box that covers all the potential configurations in the configuration space. An example of such a bounding box is shown in Figure \ref{fig:pca_with_grid}. Consider a coordinate system that originates at one of the corners of the bounding box and has its axis along the two sides of the bounding box. We will refer to one axis as $bx$ and the other as $by$. Furthermore, let one of the corners of the bounding box be the origin with coordinates $(0, 0)$. We map the origin in the controller space to the origin in the configuration space. Next, we divide the bounding box to create a 2D grid such that the sides $bx$ and $by$ are divided in the same number of increments as sliders $x$ and $y$, respectively. Each point on the configuration grid is assigned a coordinate according to the number of increments from the origin. The controller and configuration space are mapped such that a point with coordinates in the controller space $(x, y)$ is mapped to a point with coordinates $(bx, by)$ in the configuration space. Next, for each point on the grid, we determine its 2D coordinates and dimensional configuration. The configurations at the grid points constitute the presets and are included in P.

As in the case of collection-based interfaces, it is possible to evaluate and optimize the population coverage of slider-based interfaces.
To evaluate the population coverage of the slider-based approach, Algorithm \ref{alg:coverage-computation} is invoked with \presets, including all the grid points.
The population coverage depends only on the number of increments used for each of the sliders (and implicitly in the configuration grid).
We may optimize these parameters to improve the population coverage.

\section{Experiments}
\label{sec:experiments}

Our experiments answer the following questions:
\begin{itemize}
    \item What is the trade-off between the number of presets and coverage of collection- and slider-based techniques? How is this trade-off impacted by the algorithms used to determine the presets?
    \item What is the impact of user demographics (e.g., sex or age) on coverage?
    \item How robust are our results with respect to our modeling assumptions?
\end{itemize}

\subsection{Trade-off between Population Coverage and Number of Presets}

Let us start by considering the performance of the greedy and genetic algorithms in building presets for collection-based self-fitting methods. 
We also ran k-means clustering algorithm as a baseline algorithm similar to \cite{jensenCommonConfigurationsRealEar2020}.
Specifically, we ran k-means on all the preferred configurations to construct $N$ clusters. 
The presets were determined by computing the mean of all preferred configurations in the same cluster.


\Figure[t!](topskip=0pt, botskip=0pt, midskip=0pt)[scale=0.5]{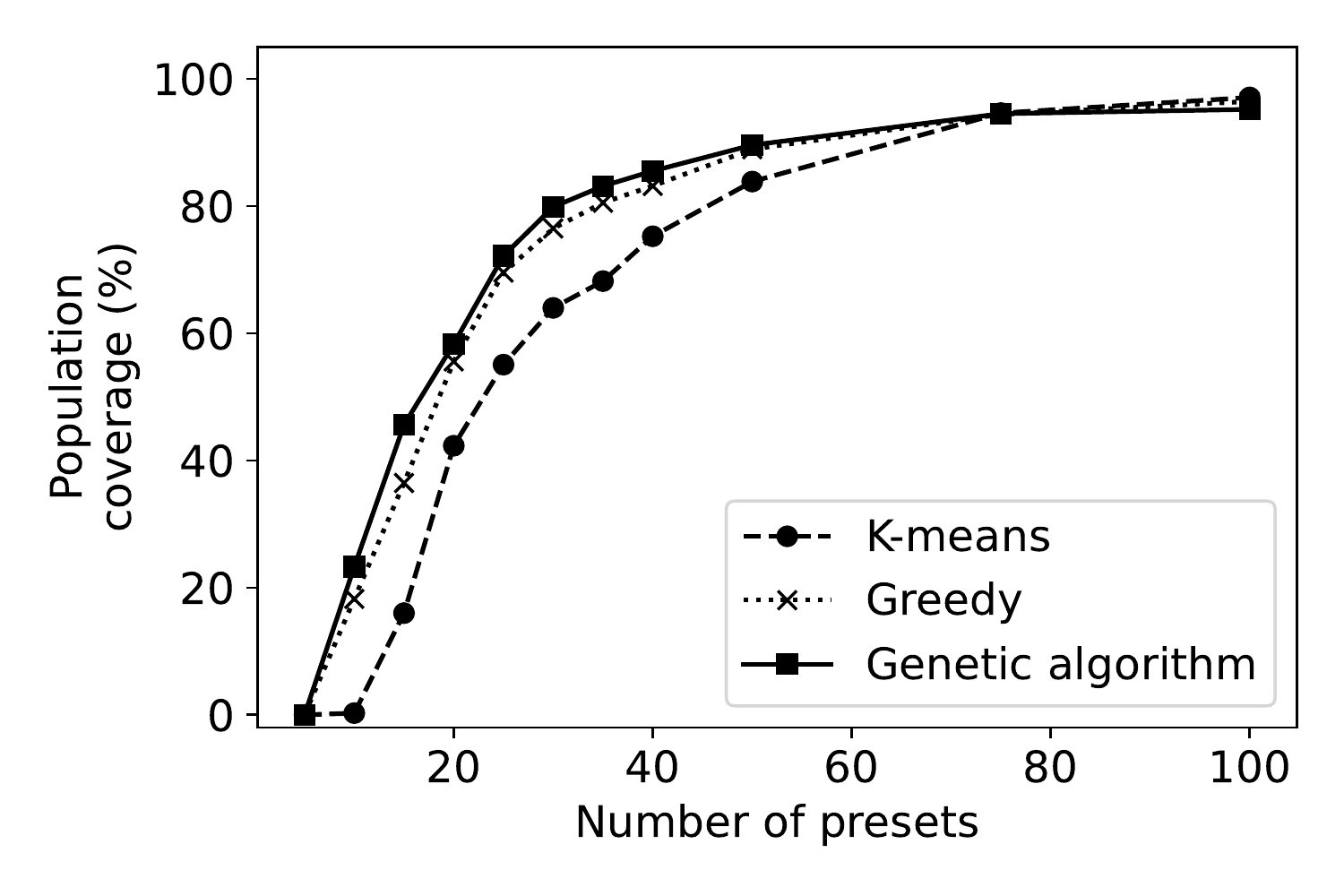}
{Population coverage for k-means, greedy, and genetic algorithm methods of determining the presets of collection-based methods.  \label{fig:coverage_data_comparison}}

Figure \ref{fig:coverage_data_comparison} plots the population coverage as the number of presets is varied from 5 to 40 in increments of 5.
As expected, the general trend is that the population coverage increases with the number of presets.
Initially, the population coverage increases fast with the number of presets.
As the number of presets continues to grow, there are diminishing returns and further improvements in population coverage diminish.
For example, in the case of the genetic algorithm, an increase from 15 to 20 presets results in a increase in coverage of 46\% to 60\%, an improvement of 14\%.
In contrast, increasing the number of presets from 35 to 40, results in an increase in population coverage of only 2.5\%.
The population coverage when the number of presets is 100 is approximately 95\%.
Improving the population coverage further would require a significant number of additional presets as the remaining 5\% of the users have configurations that differ significantly from the remainder of the population.

\noindent\textbf{Result: The population coverage of a collection-based method increases with the number presets; however the improvements diminish with additional presets. Achieving a population coverage higher than 90\% requires a large number of presets.}

The method used to derive the presets has an impact on the obtained population coverage.
The k-means clustering has consistently lower population coverage, particularly when the number of presets is below 35.
The reason behind this worse performance is that k-means is not designed to maximize population coverage.
The genetic algorithm provides a maximum of 9\% improvement in coverage over the greedy algorithm.
Note that even though a 9\% improvement may seem minor, in practice this means tens of thousands of users who might find a configuration they prefer. 
When the number of presets is large, all considered methods of generating presets provide similar performance.
Note that usually the most important cases are when the number of presets is in the range 4 -- 30, when the algorithms differ the most in their performance.
There are self-fitting methods that envision users performing simple auditory tests and being prescribed an OTC HA using one of four presets~\cite{urbanskiNewEvidenceBasedFitting2021}.
The the other end, users might be provided with an interface to select one out of 30 presets for their use.

\begin{figure*} 
\centering
\subfloat[20 Presets]{\includegraphics[width=0.32\linewidth]{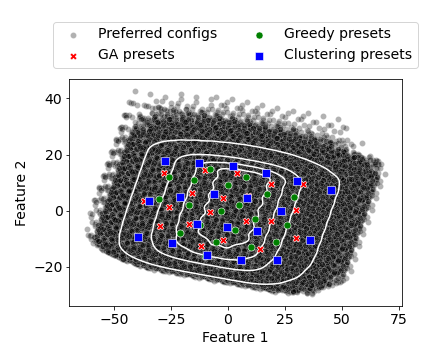}}
\hfil
\hfil
\subfloat[30 Presets]{\includegraphics[width=0.32\linewidth]{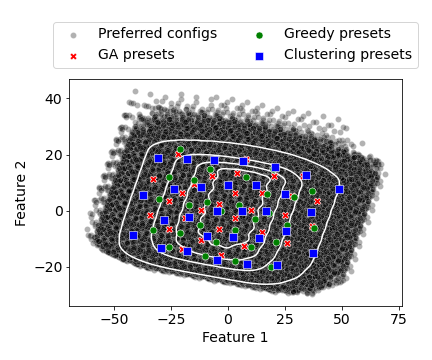}}
\hfil
\hfil
\subfloat[40 Presets]{\includegraphics[width=0.32\linewidth]{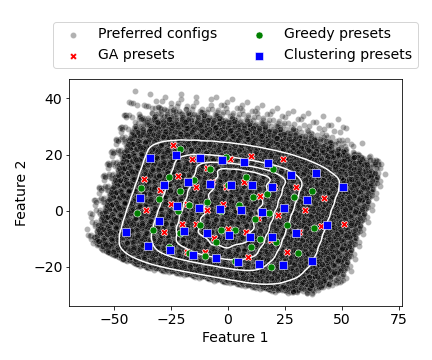}}
\caption{Representation of Presets in PCA space. (GA Presets, Greedy Presets and Clustering presets)}
\label{fig:greedy-pca-plots}
\end{figure*}

We use the greedy, genetic algorithm, and k-means to generate the 20, 30, and 40 presets.
The generated presets are projected in two dimensions using PCA and shown in Figure \ref{fig:greedy-pca-plots}.
A common trend across the presets is the k-means tends to generate presets that are more spread out than the the greedy and genetic algorithms.
As the presets picked by k-means tend to be more spread out some of them are in areas where they provide little population coverage.
In contrast, the other two methods pick that are more densely packed towards the center of the figure.
This results highlights the importance of picking an algorithm that is designed to maximize the population coverage (greedy or genetic algorithms) rather than as different objectives (for k-means).

\noindent{\textbf{Result: For the cluster-based methods that use four to twenty presets, the genetic algorithms provides better performance than the greedy and k-means approaches.}}

\Figure[t!](topskip=0pt, botskip=0pt, midskip=0pt)[scale=0.5]{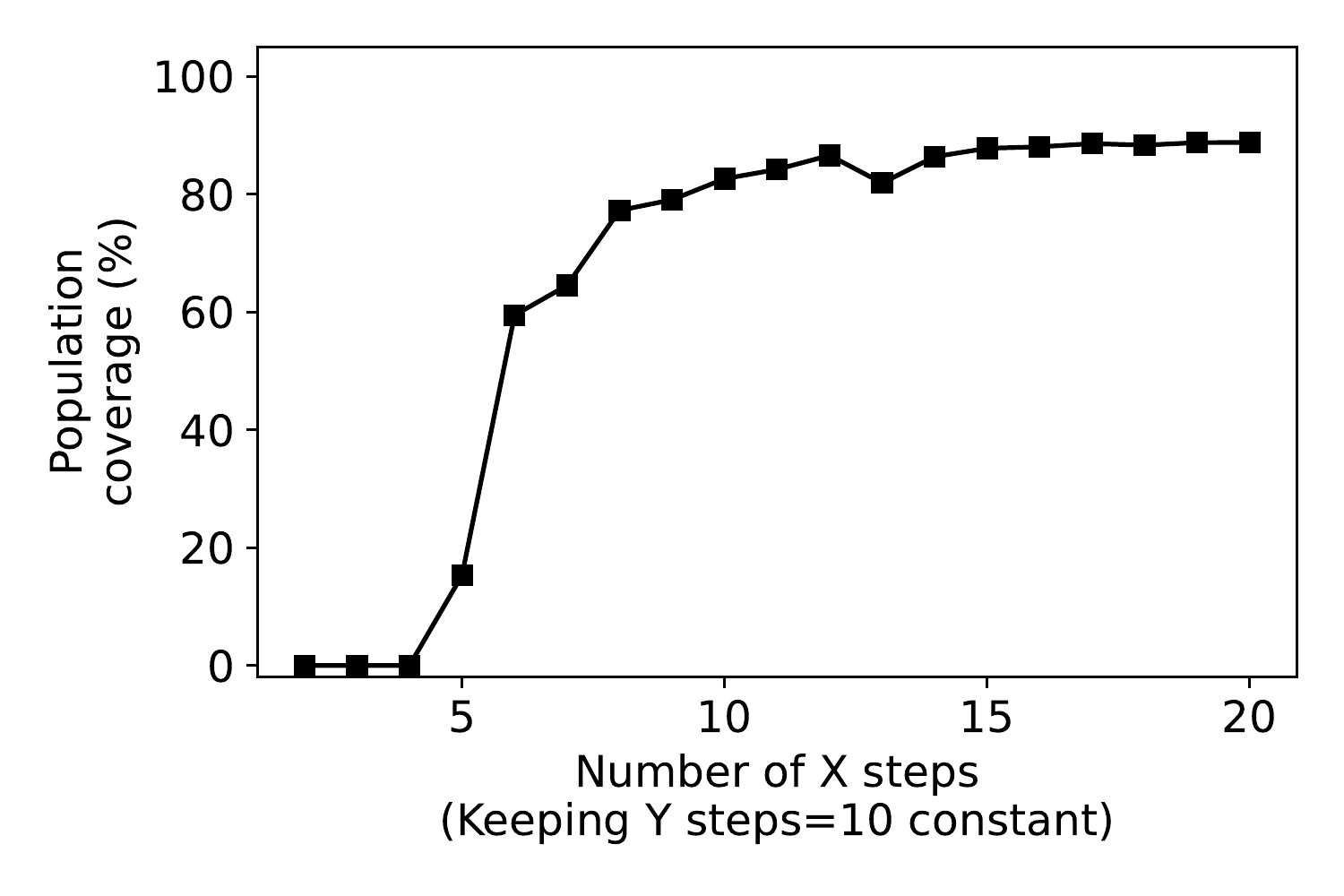}
{Percentage coverage results for slider based approach. Y steps were kept constant (10) while X steps were varied from 2 to 20.   \label{fig:slider_approach_results_varying_x_steps}}

\Figure[t!](topskip=0pt, botskip=0pt, midskip=0pt)[scale=0.5]{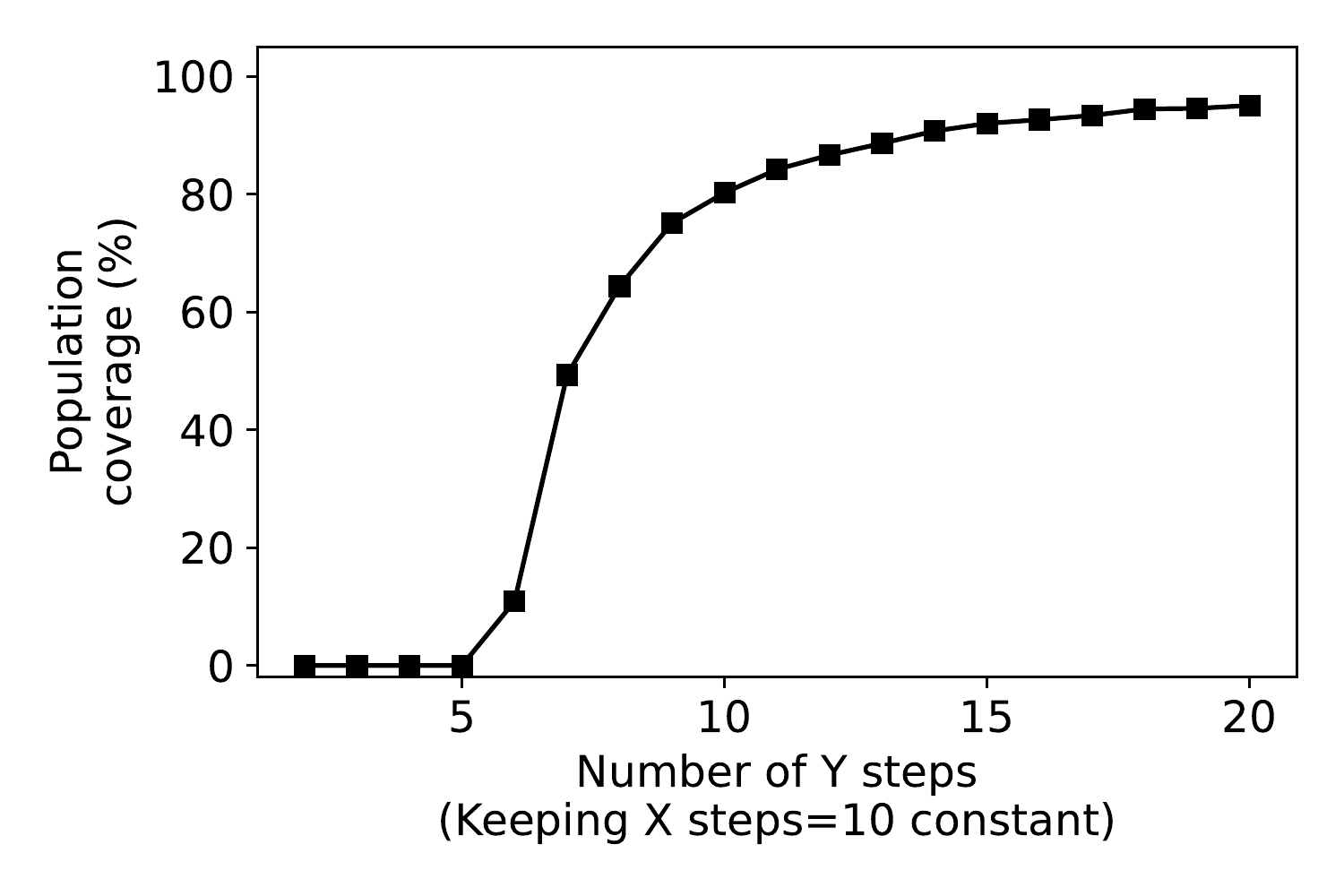}
{Percentage coverage results for slider based approach. X steps were kept constant (10) while Y steps were varied from 2 to 20.   \label{fig:slider_approach_results_varying_y_steps}}

Next, we consider user interfaces that are slider-based where the number of steps on the sliders are increased from 2 to 20.
Figure \ref{fig:slider_approach_results_varying_x_steps} and Figure \ref{fig:slider_approach_results_varying_y_steps} plot the population coverage for different number of steps used for the x and y sliders respectively while keeping y and x steps constant at 10.
As the number of steps is increased, the two-dimensional grid in the configuration space becomes denser and the associated population coverage also increases.
A user interface with the two sliders each having ten steps provides a population coverage of 78.24\%.
In general, for the same number of presets sliders provide lower coverage than the same number of presets for collection-based presets.
This is a consequence of the slider-based approaches constraining presets on the 2D grid whereas no such constraints are imposed for collection-based approaches.

\noindent\textbf{Results: The proposed approach can be used to optimize the number of increments for slider-based approaches.}


\subsection{Impact of Demographics}


One approach that may yield improvements to the population coverage is to consider the impact of various demographics. 
In the following, we will consider whether dividing the population into four subgroups based on age and gender can yield presets that are tailored for those specific groups and provide higher coverage then the original presets that do not differentiate for gender or age. The subsets into which we divide the population of users with mild-to-moderate hearing loss are:
\begin{itemize}
    \item male, age > 65
    \item male, age $\le$ 65
    \item female, age > 65
    \item female, age $\le$ 65
\end{itemize}

\begin{figure*} 
\centering
\subfloat[Subgroup: male, age > 65]{\includegraphics[width=0.45\linewidth]{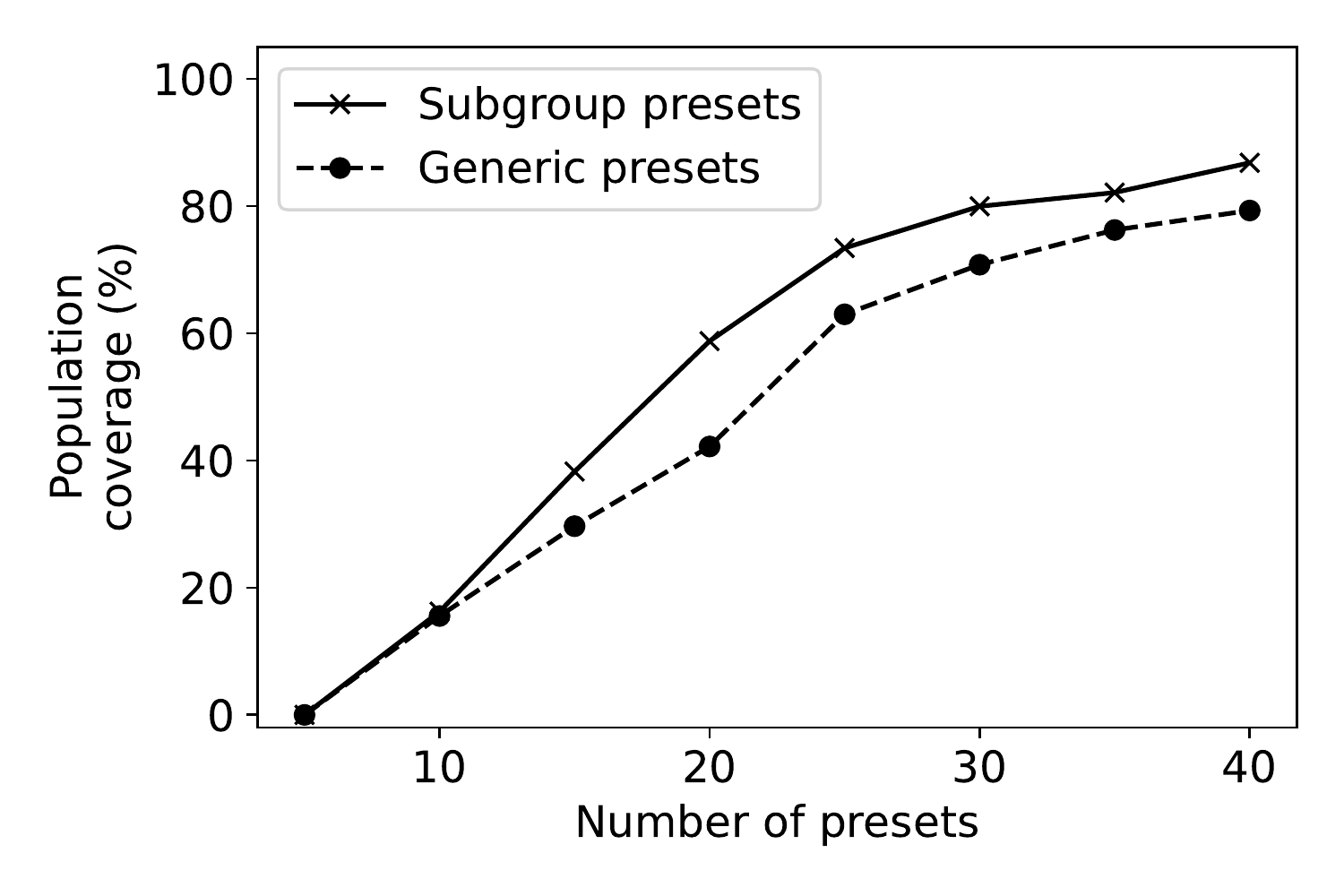}}
\hfil
\hfil
\subfloat[Subgroup: male, age <= 65]{\includegraphics[width=0.45\linewidth]{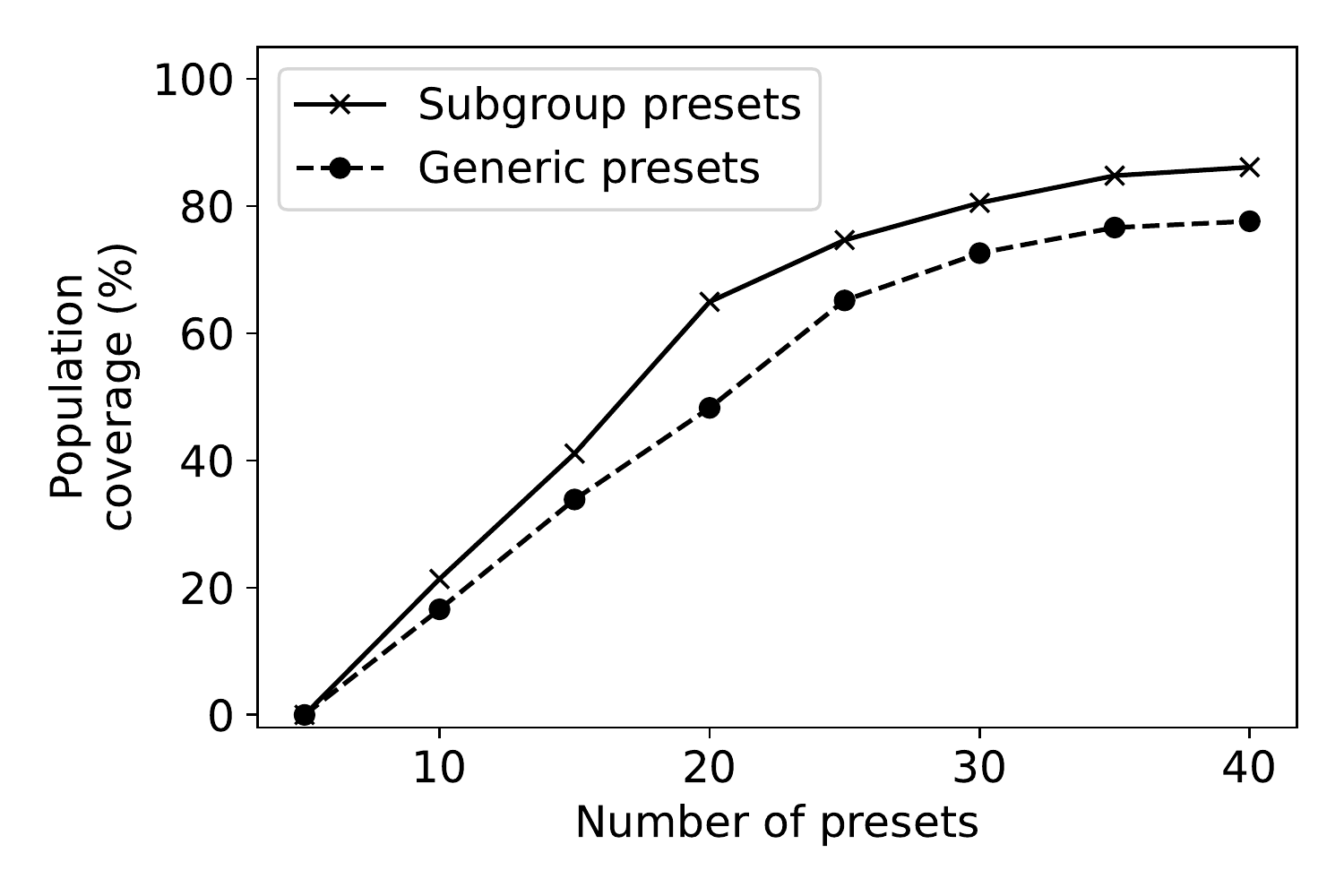}}
\hfil
\subfloat[Subgroup: female, age > 65]{\includegraphics[width=0.45\linewidth]{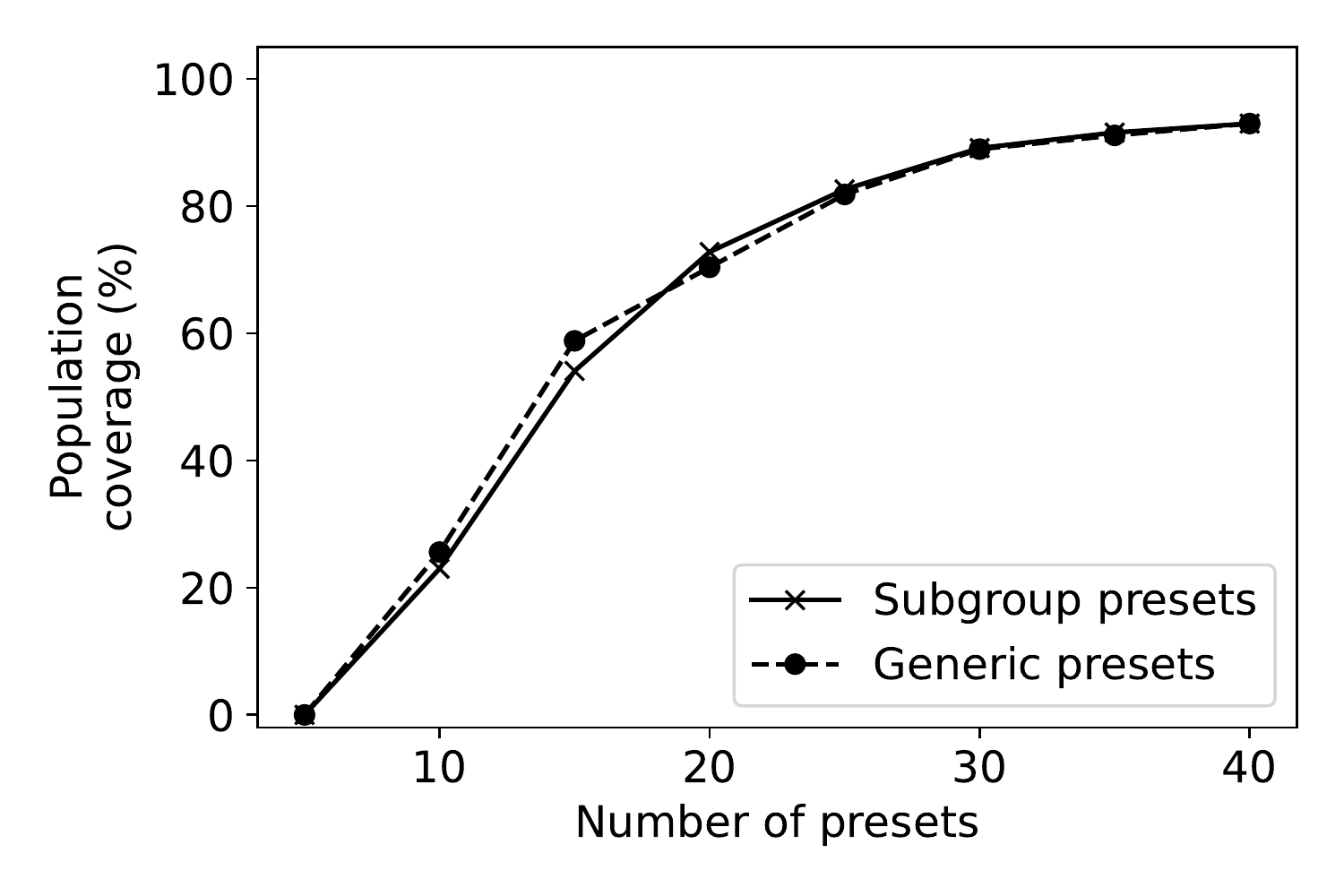}}
\hfil
\hfil
\subfloat[Subgroup: female, age <= 65]{\includegraphics[width=0.45\linewidth]{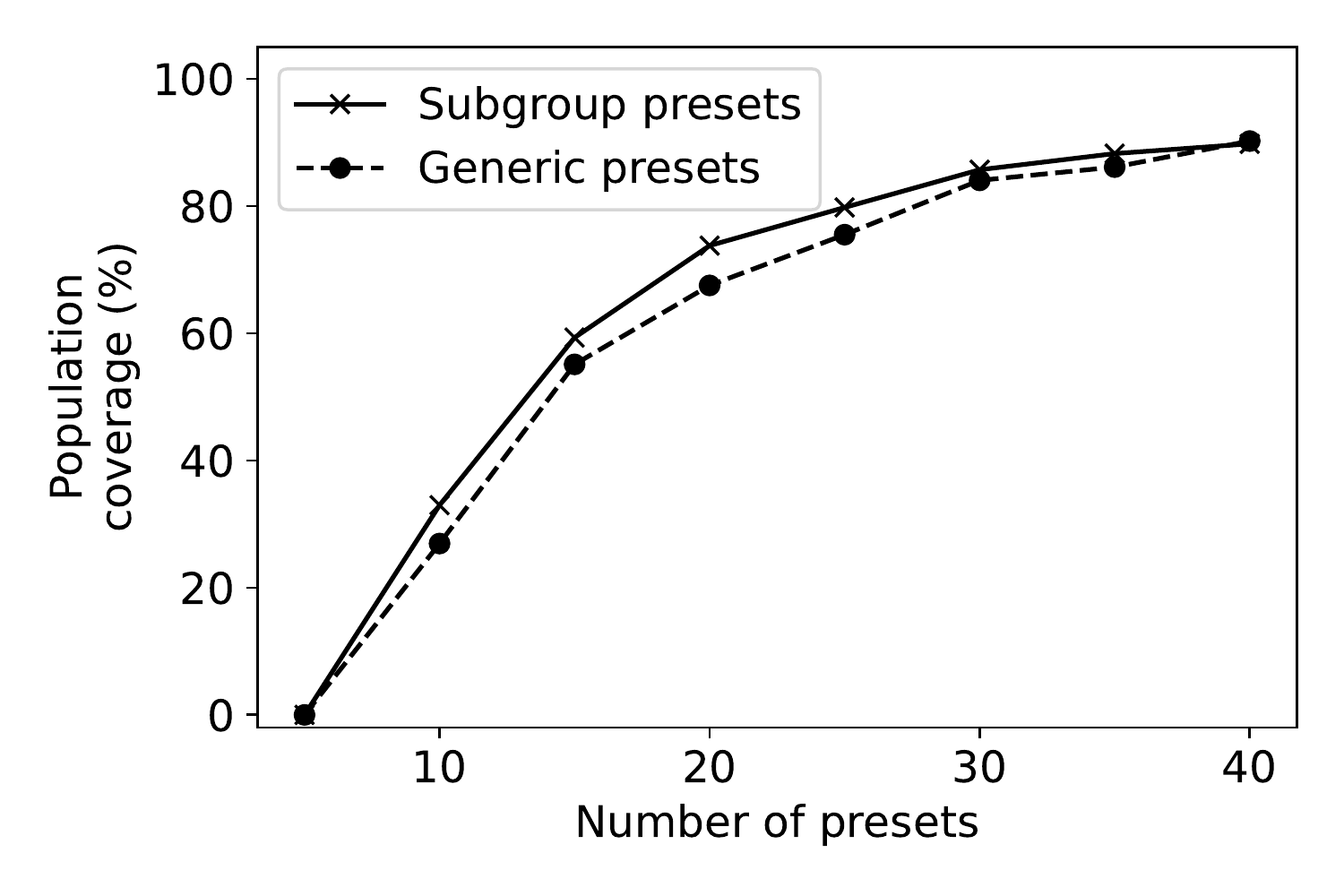}}
\hfil
\caption{Population coverage for different subgroups}
\label{fig:subgroup_results}
\end{figure*}

Figure \ref{fig:subgroup_results} plots the population coverage for each of the four groups with GA presets and presets obtained by running GA only on subgroup population.
The results indicate that in some cases higher population coverage may be achieved if the presets are build for that specific subset of the population.
The most promising results are for the subgroup which includes males who are over 65.
In this case, constructing presets specialized for this subgroup provide an improvement of 16\% in population coverage over using the presets constructed for the general population.
However, these results do not hold for all subgroups. 
Our approach provides no meaningful improvements for the subgroup which included women over 65 over constructing presets for the overall population.

\noindent\textbf{Result: Using demographics to refine the target population of users may yield significant improvements in population coverage.}



\subsection{Robustness of Results}

\Figure[t!](topskip=0pt, botskip=0pt, midskip=0pt)[scale=0.5]{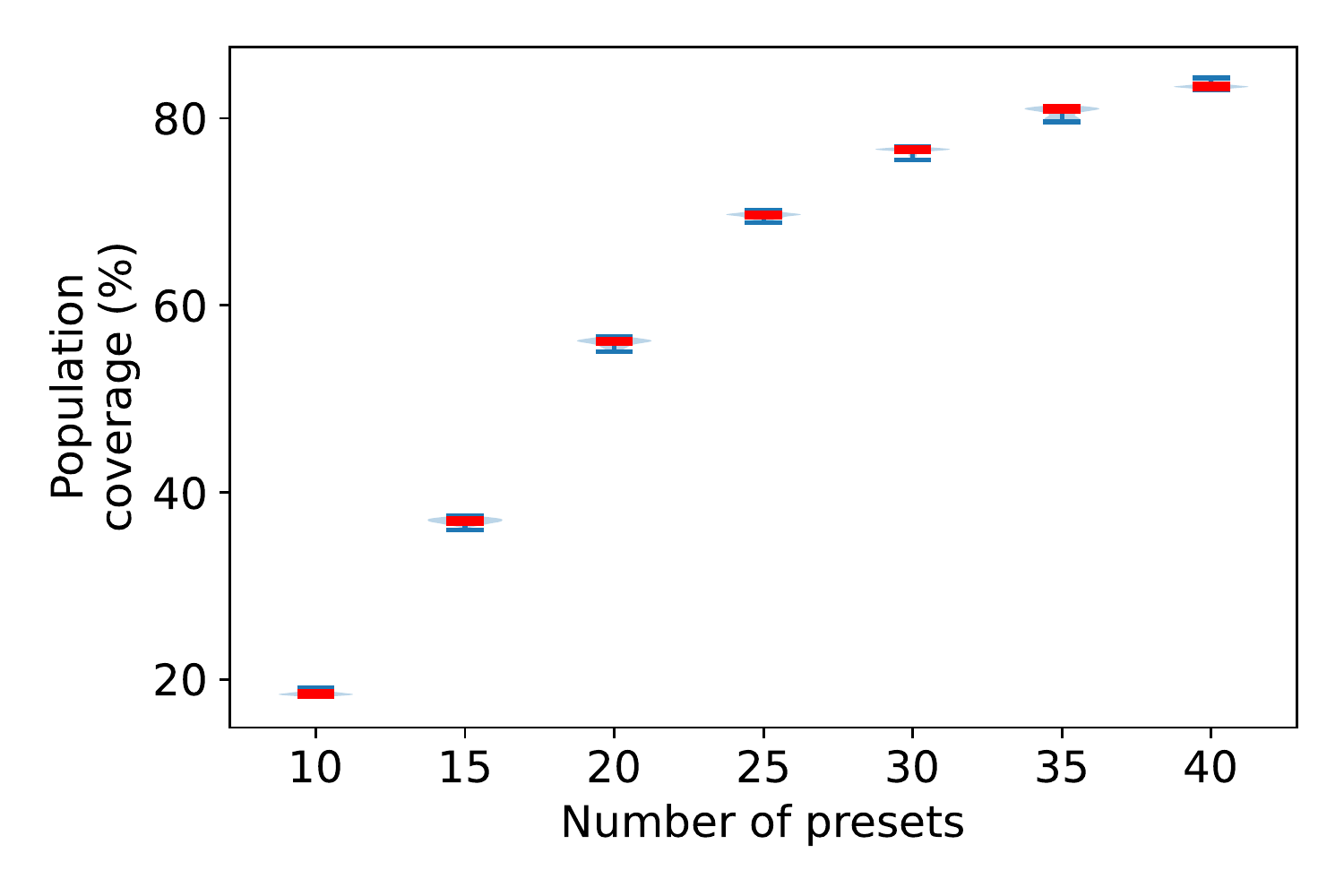}
{Distribution of population coverage by greedy presets using bootstrap technique  \label{fig:boostrap_greedy_results}}

An important question is to evaluate how robust are our methods to different ways of characterizing a user's deviation from NAL-NL2.
We will use two different approaches -- using bootstrap and scaling the variance of the Gaussian.
We used bootstrap~\cite{boothroydAmplificationSelfAdjustmentControls2022, mackersieHearingAidSelfAdjustment2020} to obtain a distribution of possible values for the variance of the 2D Gaussian.
Accordingly, we sample with replacement from the empirical data of studies ~\cite{boothroydAmplificationSelfAdjustmentControls2022, mackersieHearingAidSelfAdjustment2020} to generate 50 bootstrap samples.
For each bootstrap sample, we run the greedy algorithm for 5 to 40 presets and compute their associated population coverage.
Figure \ref{fig:boostrap_greedy_results} plots the probability density function of the population coverage for the 20, 25, 30, 35 and 40 presets.

The figure shows that there is little variation in the population coverage for the generated bootstrap samples.
Moreover, the variation tends to diminish with the number of presets.
At 20 presets, population coverage range fluctuates by $\pm$ 0.75\%.
With increasing number of presets, this variation in range significantly goes down.
At 40 presets, fluctuation range is $\pm$ 0.25\%.
Therefore, there are small variations in the population coverage within the bootstrap samples.
This indicates that if the data from original studies used to build the 2D Gaussian is representative, then we should observe little variation in the population coverage.

Another approach to evaluate the robustness of our approach is to consider the impact that a higher Gaussian variance may have.
We have scaled the fitted variance using the data from studies \cite{boothroydAmplificationSelfAdjustmentControls2022, mackersieHearingAidSelfAdjustment2020} by scalars 0.5, 1, and 1.5.
The expectation is that increasing the variance will result in lower coverage for the same number of presets.
Stated differently, a larger number of presets is necessary to achieve the same coverage.
We have evaluated the impact of scaling the variance on the case of using the greedy algorithm to build presets.

\Figure[t!](topskip=0pt, botskip=0pt, midskip=0pt)[scale=0.5]{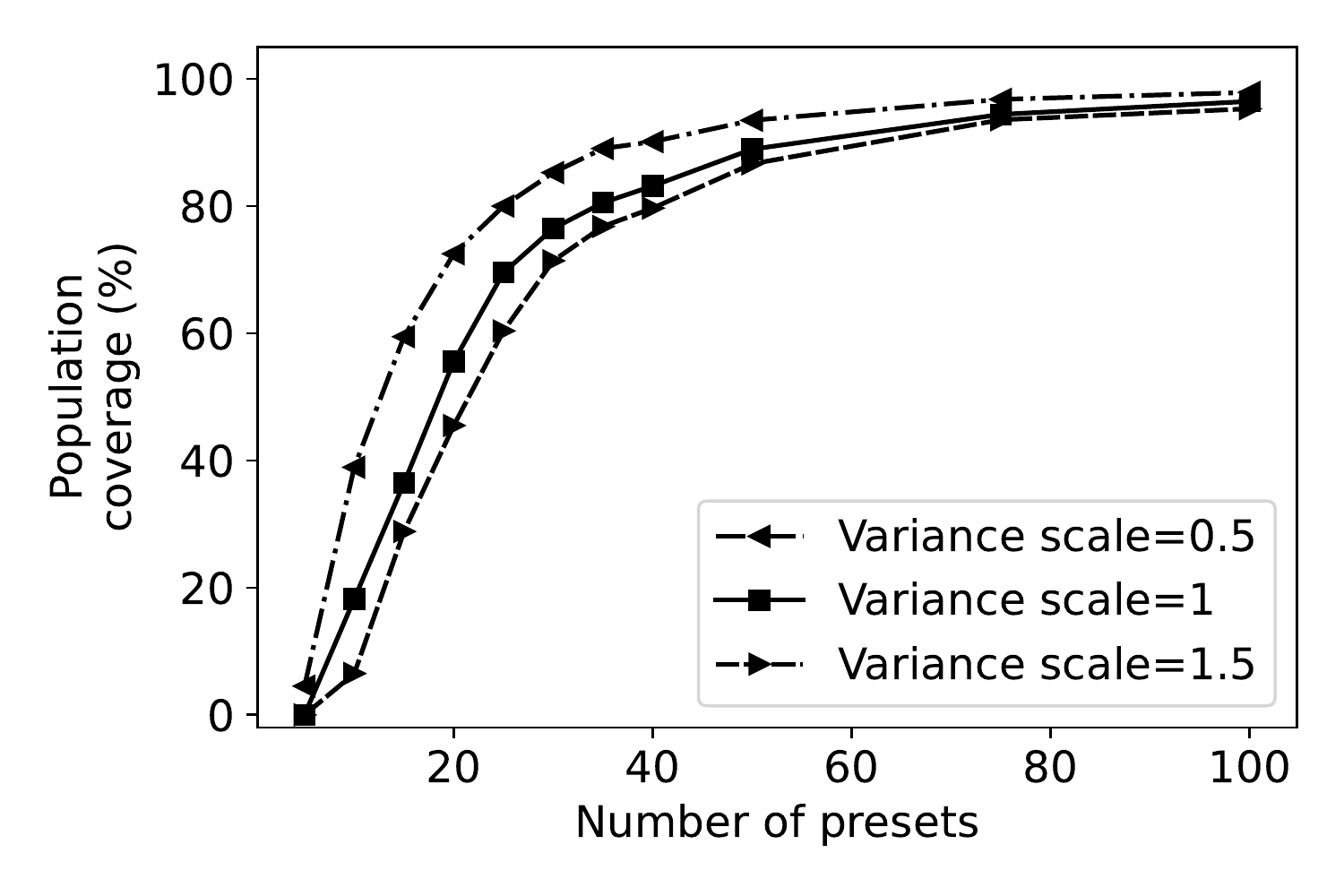}
{Effects of variance scaling on greedy preset coverage  \label{fig:variance_scaling_greedy}}

Figure \ref{fig:variance_scaling_greedy} plots the effects of variance scaling on population coverage by greedy presets.
When the variance is 1.5 times larger, there is a reduction of 3.5\% in population coverage.
Conversely, when the variance is 0.5 times smaller, there is an increase of 7\% in population coverage at 40 presets.
Overall, these results indicate that population coverage results are robust with respect to larger variances that those expected based on the empirical data alone.

Intuitively, if variance is lower, there is a higher variation weight associated with the variations closer to user's REIG.
Thus, it takes less number of presets to cover the same population.
It reflects in our results i.e., population coverage difference is as high as 32\% for 10 greedy presets generated between variance scale 0.5 and 1.5.
This wide difference narrows down to almost 7\% for 50 greedy presets.

\noindent\textbf{Result: The experiments demonstrate that the population coverage results are robust different assumptions about the underlying Gaussian distribution controlling the weights of transfer functions.}

\section{Discussion}
\label{sec:discussion}

The paper uses population coverage to evaluate and optimize various settings for preset-based methods.
We caution the reader that population coverage not the only metric that may be used to assess self-fitting methods.
To fully evaluate self-fitting methods, it is necessary to run detailed user studies to evaluated their performance.
However, population coverage can guide how user studies may be set up.
For example, in the case of collection-based approaches, the region of interest is when the number of presets are in the range of 20 -- 40.
Using fewer than 20 presets yields would cover only a small fraction of users.
In contrast, increasing the number of presets beyond 40 results in minor improvements on population coverage.
Further user studies should particularly focus in this range. 
Similarly, in the case of slider-based approaches, the region of interest is when the number of increments is about ten for both sliders.


The computation of population coverage depends on a number of modeling assumptions and hyper-parameters.
The hyper-parameters of our model include $\gamma$, $R$, and likelihood of each possible configuration as given by the two dimensional Gaussian.
Our experiments show that the population coverage values are fairly robust to increases in the variance of the Gaussian distribution.
The parameters $\gamma$ and $R$ are configured best on audiology expertise, however, other values would be reasonable too.
Our experimentation with different values for $\gamma$ and $R$ shows that even though population coverage changes, the same overall trends are observed.

\section{Conclusion}
\label{sec:conclusions}



This paper proposes a novel metric -- population coverage -- to evaluate, compare, and optimize preset-based self-fitting methods.
The population coverage estimates how well a given number of presets meet the needs of populations of users with different characteristics.
The unique aspect of our approach is that the population coverage is computed to account for how a user's preferred configuration may deviate from the NAL-NL2 prescription.
We apply the population coverage to optimize the presets used by collection- and slider-based methods.
Specifically, greedy and genetic algorithms are proposed presets that maximize population coverage.
Similarly, algorithms are proposed to configure the number of intervals on slider-based interfaces to maximize coverage.
Our experiments indicate that the proposed algorithms can effectively identify presets for both collection- and slider-based methods.
Moreover, we may use population coverage to narrow down the different configurations of preset-based methods with good population coverage and whose performance should be further characterized through user studies.

\bibliographystyle{IEEEtran}
\bibliography{ref}

\EOD

\end{document}